\documentclass[lettersize,journal]{IEEEtran}
\usepackage{amsmath,amsfonts}
\usepackage{algorithmic}
\usepackage{algorithm}
\usepackage{array}
\usepackage[caption=false,font=normalsize,labelfont=sf,textfont=sf]{subfig}
\usepackage{textcomp}
\usepackage{stfloats}
\usepackage{url}
\usepackage{verbatim}
\usepackage{graphicx}
\usepackage{cite}
\usepackage{booktabs}   
\usepackage{multirow}   
\usepackage{makecell}   
\usepackage{adjustbox}
\usepackage{amssymb}
\usepackage{hyperref}   
\usepackage{tabularx}
\usepackage[normalem]{ulem}  
\usepackage{float}
\usepackage{svg}
\usepackage{algorithm}
\usepackage{algorithmic}
\usepackage{listings}

\lstdefinelanguage{json}{
    basicstyle=\ttfamily\footnotesize,
    numbers=left,
    numberstyle=\scriptsize,
    numbersep=6pt,
    stepnumber=1,
    frame=none,
    showstringspaces=false,
    breaklines=true,
    breakatwhitespace=false,
    columns=fullflexible,
    keepspaces=true,
    tabsize=2
}
\begin{document}

\title{Di²CycleSB: Towards High-Quality Unsupervised Nighttime Visibility Enhancement via Schr\"odinger Bridge Transformer}

\author{Hanting Li, Xin Sun,~\IEEEmembership{Senior Member,~IEEE,} Fei Zhou, Wei Ye, Jungong Han,~\IEEEmembership{Senior Member,~IEEE,} Liang-jie Zhang,~\IEEEmembership{Fellow,~IEEE}
\thanks{This work is supported by the Ministry of Science and Technology- National Key Research and Development Program of China (2025YFE0202900), Science and Technology Development Fund (FDCT-MOST), Macao SAR (0007/2025/AMJ), Science and Technology Development Fund - International Collaborative Research, Macao SAR (0001/2025/AIJ), Science and Technology Development Fund, Macao SAR - Basic Research (0006/2024/RIA1), and National Natural Science Foundation of China (62441235)} 
\thanks{H. Li and X. Sun are with Faculty of Data Science, City University of Macau, 999078, SAR Macao, China. F. Zhou is with College of Oceanography and Space Informatics, China University of Petroleum (East China), Qingdao, China. W. Ye is with College of Electronic and Information Engineering, Shanghai Research Institute for Autonomous Intelligent Systems, State Key Laboratory of Autonomous Intelligent Unmanned Systems, Tongji University, Shanghai, China. J. Han is with Department of Automation, Tsinghua University, Beijing, China. L.J. Zhang is with College of Computer Science and Software Engineering, Shenzhen University, Shenzhen, China.}

}

\markboth{Journal of \LaTeX\ Class Files,~Vol.~14, No.~8, August~2021}%
{Shell \MakeLowercase{\textit{et al.}}: A Sample Article Using IEEEtran.cls for IEEE Journals}

\IEEEpubid{0000--0000/00\$00.00~\copyright~2021 IEEE}

\maketitle

\begin{abstract}
  Light-effect contamination poses a significant challenge to nighttime visibility enhancement. Most methods suppress light effects by estimating and decomposing them through prior-driven regularization, yet they are often limited by hand-crafted priors and ill-posed nature of decomposition. This work proposes Di²CycleSB, a unsupervised Cycle Schr\"odinger Bridge Transformer framework guided by dynamic integral image priors, for high-quality unsupervised nighttime visibility enhancement. Specifically, a novel light-effect estimator is introduced to parameterize Gaussian-like adaptive priors by aggregating dynamic integral image representations for non-uniform glow estimation. Then, we propose a prior-informed Generator that exploits light-effect representations to guide long-range dependency modeling within our specific Transformer blocks. We formulate light-effect suppression as a Schr\"odinger bridge problem and construct forward and backward bridges with cycle consistency constraints to achieve visually pleasing enhancement. Extensive experiments on real-world datasets demonstrate the remarkable effectiveness of our Di²CycleSB in enhancing nighttime visibility. In particular, it achieves effective end-to-end light-effect suppression without any regularization constraints and image decomposition. The code and models are available at \url{https://github.com/LHTcode/Di2CycleSB}.
\end{abstract}


\vspace{-10pt}
\section{Introduction}
\IEEEPARstart{D}{ue} to non-uniform illumination conditions and multiple light sources, night images often suffer from locally insufficient visibility and light-effect contamination (e.g., glare and floodlight artifacts). Most existing low-light image enhancement methods~\cite{jiang2021enlightengan,11125898,11574000} overlook the suppression of light effects and even tend to amplify light-effect contamination. Some nighttime dehazing methods~\cite{zhang2020nighttime,wang2022variational} can alleviate glow artifacts. However, they typically enforce global haze suppression, which often results in over-cleaned and visually unnatural appearances.

\begin{figure}[!t]
        \centering
        \includegraphics[width=\linewidth]{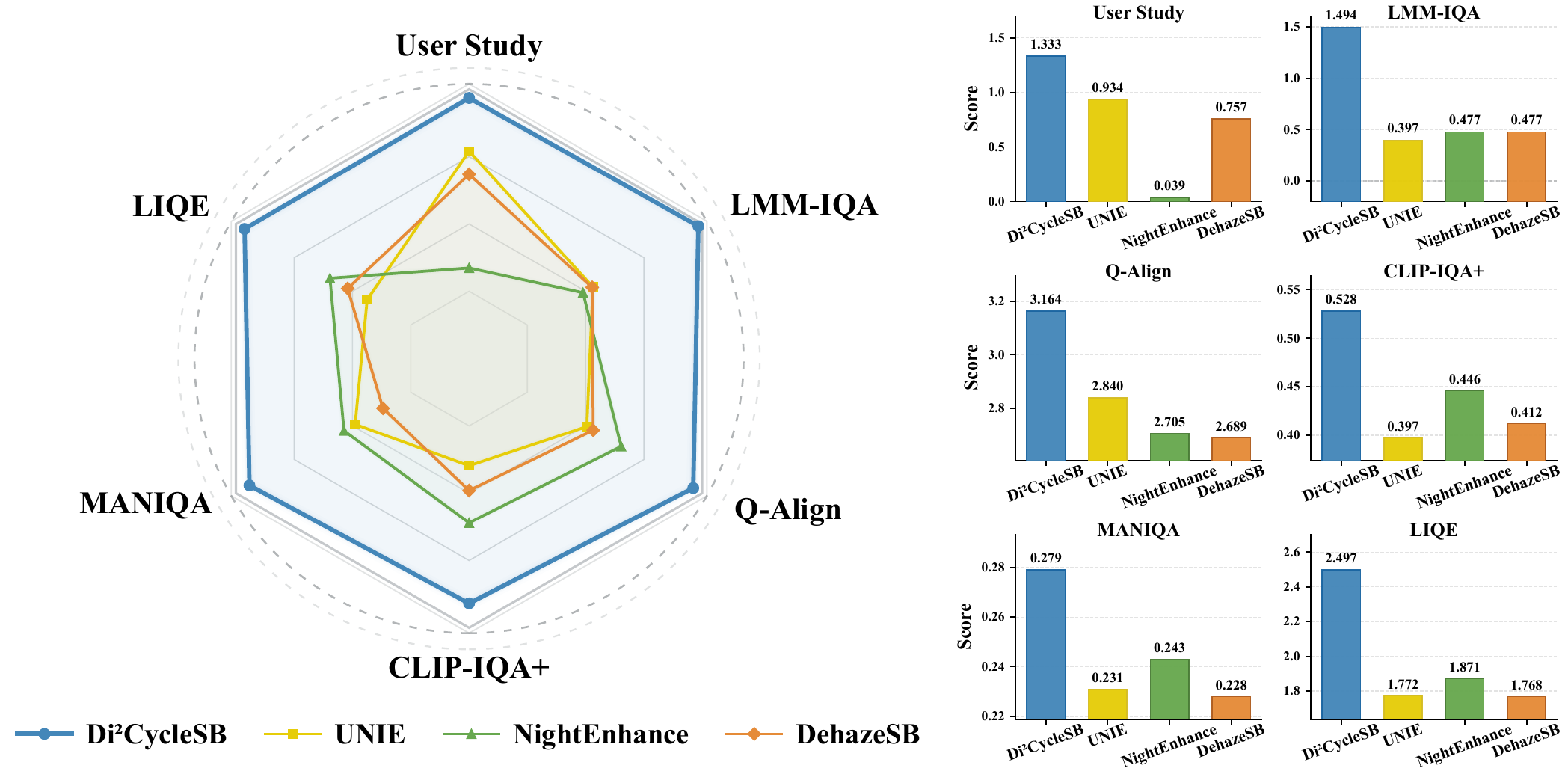}
        \vspace{-18pt}
        \caption{Comparison of Di²CycleSB, UNIE~\cite{jin2022unsupervised}, NightEnhance~\cite{jin2023enhancing}, and DehazeSB~\cite{lan2025schrodinger} on the Light-Effects~\cite{sharma2021nighttime} and NTIRE 2025 Challenge~\cite{Ershov_2025_CVPR} datasets. Our method achieves the best performance across \hyperref[sec:user_study]{User Study}, \hyperref[sec:LMM]{LMM-IQA}, Q-Align~\cite{wu2024qalign}, CLIP-IQA+~\cite{wang2023clipiqa}, MANIQA~\cite{yang2022maniqa}, and LIQE~\cite{zhang2023liqe}. For visualization, each metric is first normalized separately for each dataset and then averaged.}
        \label{fig:first_pic}
        \vspace{-5pt}
\end{figure}

\begin{figure*}
    \vspace{-18pt}
    \captionsetup[subfigure]{labelsep=none,format=plain,labelformat=empty,labelfont=footnotesize,textfont=footnotesize}
    \centering
    \subfloat[Input]{\includegraphics[width=0.245\linewidth]{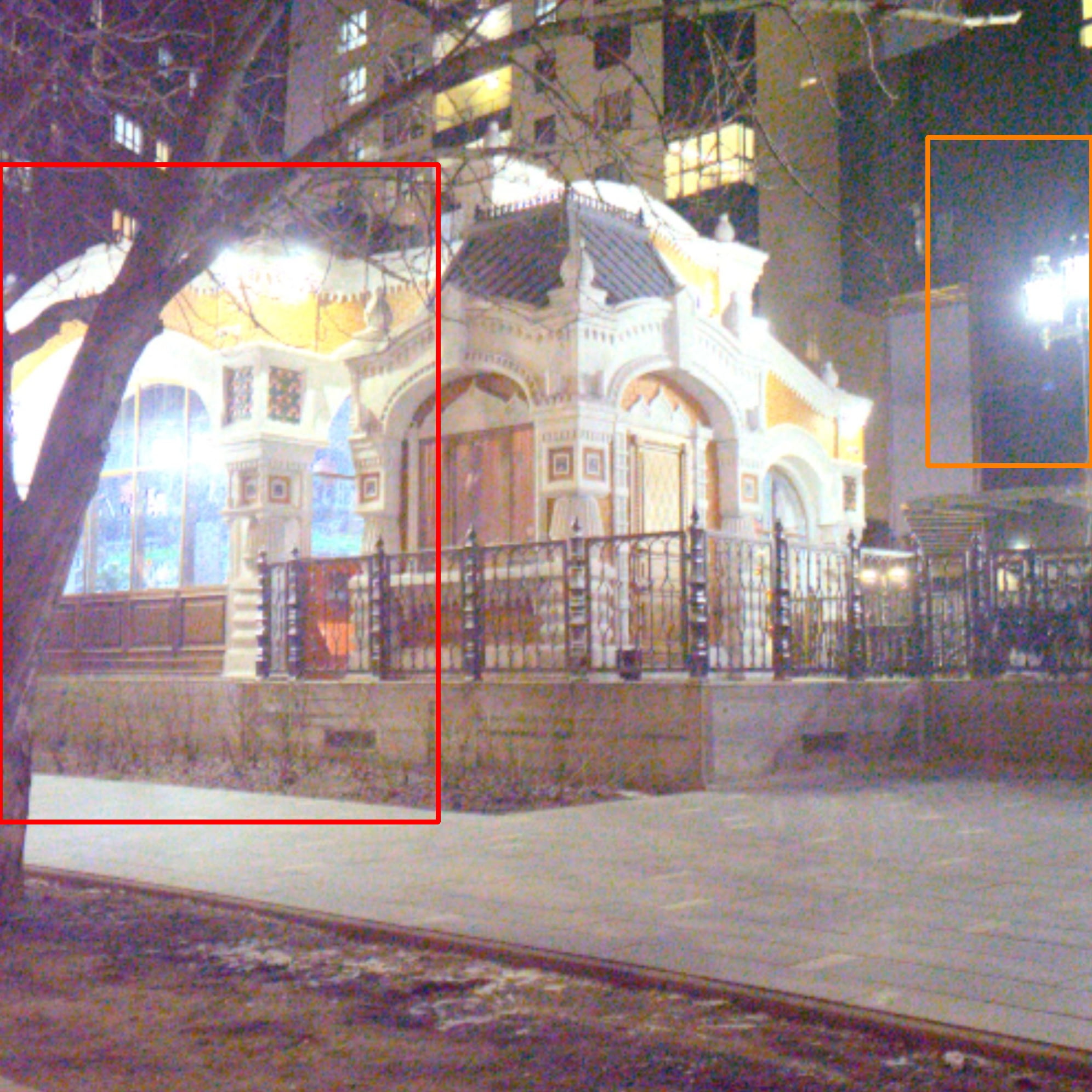}}
    \hfill
    \subfloat[SFSNiD~\cite{cong2024semi}]{\includegraphics[width=0.245\linewidth]{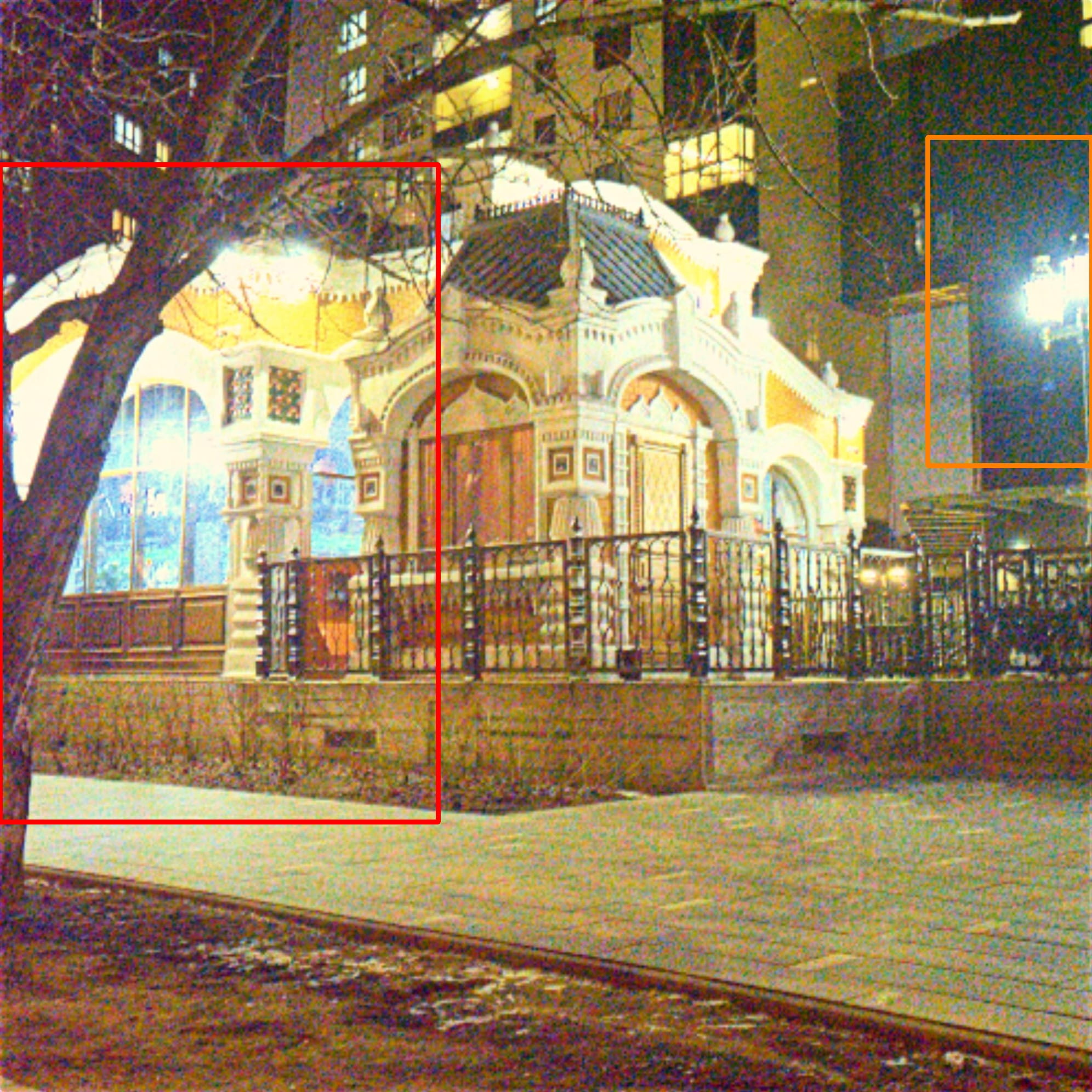}}
    \hfill
    \subfloat[UNIE~\cite{jin2022unsupervised}]{\includegraphics[width=0.245\linewidth]{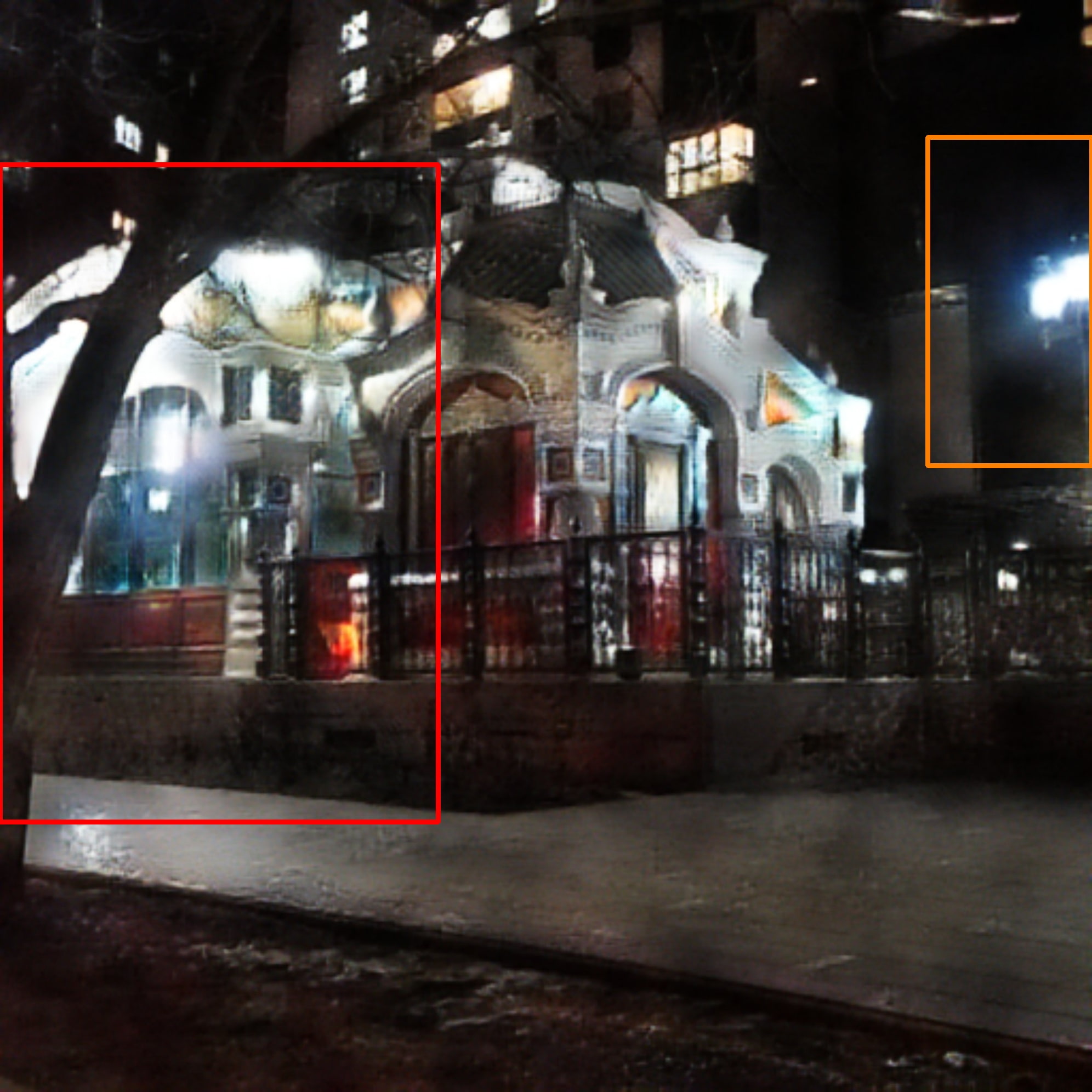}}
    \hfill
    \subfloat[Di²CycleSB (Ours)]{\includegraphics[width=0.245\linewidth]{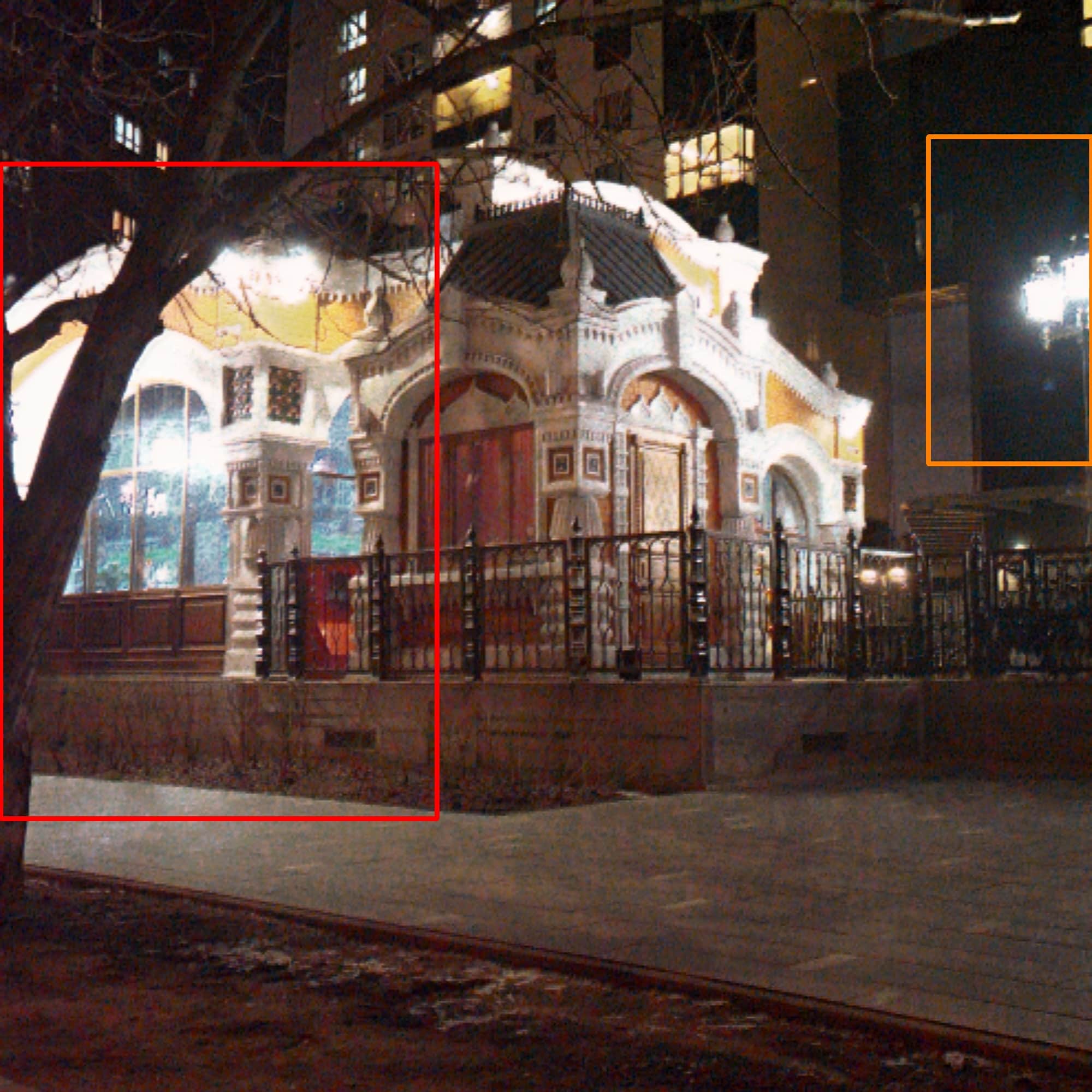}}
    \caption{Our results lead to a superior nighttime visibility experience. Di²CycleSB naturally suppresses light effects while preserving visual fidelity. All unsupervised and semi-supervised methods methods are retrained for a fair comparison. Please zoom in for better viewing.}
    \label{fig:sec_fig}
    \vspace{-15pt}
\end{figure*}

\IEEEpubidadjcol
Several recent studies have focused on light-effect suppression and visibility enhancement for night images. Sharma et al.~\cite{sharma2021nighttime} proposed the first work on light-effect suppression, which subtracts light effects from the low-frequency image component under complex constraints of color constancy, light-effect smoothness, and reconstruction. UNIE~\cite{jin2022unsupervised} formulates light-effect suppression as a layer decomposition problem which assumes that the light-effect layer follows a short-tailed distribution. And it employs a CNN-based single-generator Generative Adversarial Network (GAN) framework with unpaired training to enhance light-effect suppression. Semi-supervised methods, such as SFSNiD ~\cite{cong2024semi}, rely on pseudo-labels generated from models pretrained on synthetic datasets. Consequently, they fail to effectively mitigate diverse lighting variations in real-world scenes. Moreover, some methods model light-effect formation with the Atmospheric Point Spread Function (APSF)~\cite{wu2023generation,wu2024overall}. Overall, existing light-effect suppression methods rely on ill-posed decompositions under an additive light-effect modeling assumption which depends on either prior-based losses or hand-crafted prior guidance. Therefore, these approaches struggle to accurately suppress complex light effects in challenging scenes, often leading to incomplete suppression or noticeable artifacts, as illustrated in Fig.~\ref{fig:sec_fig}. In addition, existing unpaired training approaches typically utilize GANs with a single CNN-based generator. The potential mode collapse issue in GANs further complicates the optimization process, while the inherent incapacity of CNNs to capture long-range dependencies often leads to insufficient light-effect suppression.

Accurately modeling and suppressing non-uniform light effects is a challenging problem. Our key insights into addressing this challenge are as follows: \textbf{1)} the low-frequency prior of glow should be embedded into the network architecture; \textbf{2)} the glow prior should be learnable and spatially adaptive rather than fixed; and \textbf{3)} the unsupervised framework should be intrinsically prior-informed. Therefore, we propose a novel nighttime visibility enhancement framework, i.e., \textbf{D}ynamic-\textbf{i}ntegral-\textbf{i}mage-prior-guided \textbf{Cycle} \textbf{S}chr\"odinger \textbf{B}ridge transformer (Di²CycleSB). Our Di²CycleSB achieves end-to-end effective light-effect suppression \textit{\textbf{without}} complex hand-crafted priors, \textit{\textbf{without}} any specialized training strategies, and \textit{\textbf{without}} image decomposition. Specifically, we formulate the transformation between light-effect-affected and light-effect-free image distributions as a Schr\"odinger bridge problem, which is solved via iterative adversarial learning. Furthermore, during training, we construct forward and backward Schr\"odinger bridges and impose cycle-consistency constraints between the forward and backward processes at arbitrary time steps. To embed spatially variant low-frequency priors, we propose a \textbf{D}ynamic-\textbf{i}ntegral-\textbf{i}mage-based \textbf{L}ight-effects \textbf{E}stimator (Di²LE). The kernel is a dynamic integral image algorithm which can be interpreted as a low-pass filtering operator with an adaptive receptive field. Together with learnable aggregation weights, Di²LE parameterizes Gaussian-like priors for estimating non-uniform light effects. In addition, we propose a \textbf{P}rior-\textbf{I}nformed Generator (PI Generator), which leverages the estimated light effects at multi-scales to guide Transformer blocks in modeling non-local interactions among regions with various light-effect conditions. 

It is noteworthy that we are the first to achieve light-effect suppression on high-resolution unpaired night images. This is attributed to the adaptive light-effect estimation capability and cycle multi-step models of Di²CycleSB for solving the Schr\"odinger bridge, which pave the way for high-quality nighttime visibility enhancement. Overall, our contributions can be summarized as follows:

\begin{itemize}
\item We propose a novel framework, \textbf{Di²CycleSB}, which introduces the Schr\"odinger bridge into nighttime visibility enhancement. To the best of our knowledge, this is the first unpaired Cycle Schr\"odinger Bridge framework, enabling effective light-effect suppression without relying on any form of ill-posed image decomposition.
\item We present a novel light-effect estimator, \textbf{Di²LE}, which achieves spatially adaptive light-effect estimation by aggregating responses of dynamic low-pass integral-image operators to parameterize Gaussian-like priors, without loss-function constraints.
\item We propose the \textbf{PI Generator}, which leverages the estimated light effects to guide long-range dependency modeling for end-to-end light-effect suppression. To the best of our knowledge, this is the first Transformer-based unpaired Schr\"odinger bridge.
\item Extensive experiments demonstrate that our method substantially improves the visual perception and image quality of night images, and significantly outperforms existing approaches.
\end{itemize}

The rest of this paper is organized as follows. Section~\ref{sec:related-work} reviews related work in nighttime light effects suppression and Schr\"odinger bridge based methods. Section~\ref{sec:method} introduces our proposed Di²CycleSB framework and its specific modules: Di²LE, PI Generator and LLIE module. Section~\ref{sec:experiment} presents both quantitative and qualitative experimental results on various LLIE datasets. Finally, Section~\ref{sec:conclusion} concludes the proposed method and its contributions.

\vspace{-12pt}
\section{Related Work}
\label{sec:related-work}
\subsection{Nighttime Light Effects Suppression} 
\noindent Nighttime scenes contain various light sources, from which light rays are scattered multiple times in the atmospheric medium and propagate to the camera sensor from different directions with varying intensities~\cite{1211417}. Consequently, glow effects are formed on the image plane, and these adverse glows degrade nighttime scene visibility. Therefore, suppressing light effects is critical for improving the visibility of night images. Although existing low-light image enhancement methods~\cite{jiang2021enlightengan,10814698} can improve the illumination of dark regions in night images, they neglect the suppression of light effects. As a result, these methods typically amplify light effects in night images, instead further impairing visibility.

Sharma et al.~\cite{sharma2021nighttime} were the first to focus on suppressing light effects. It employs a pipeline composed of camera response function (CRF) estimation, high-low frequency decomposition, and HDR imaging to suppress light effects and enhance the dynamic range of night images. However, this method still relies on paired data for semi-supervised training. UNIE~\cite{jin2022unsupervised} estimates light effects through image layer decomposition network and develops a unsupervised light-effect-guided suppression network based on a GAN framework. However, the layer-decomposition-based approaches rely on multiple prior-driven loss constraints, making it difficult to suppress light effects in complex real-world scenes. Several methods model light effects with the atmospheric point spread function (APSF)~\cite{wu2023generation,wu2024overall}. They are inherently limited by applying globally consistent light-effects estimation method and rely on image decomposition, which is a ill-posed problem. 

Recently, several paired datasets for light-effect suppression have been introduced. Flare7K~\cite{dai2022flare7k} provides synthetic flare images, while Flare7K++~\cite{10541091} extends it with real-captured flares and light-source annotations. NightLight~\cite{xie2025learning} further incorporates both low-light conditions and light effects. However, synthetic data suffer from limited generalizability, and collecting real-world paired data remains challenging, making unpaired learning a promising alternative.

Like clear nighttime images, hazy nighttime images also suffer from severe light effects around active light sources. Several recent nighttime dehazing methods have focused on suppressing light effects in hazy nighttime scenes. Zhang et al.~\cite{zhang2020nighttime} introduced an optimal-scale maximum reflectance prior for color correction and haze removal, while Fang et al.~\cite{wang2022variational} employed a gray haze-line prior to separate glow and haze. NightEnhance~\cite{wu2024overall} adopts APSF-guided glow rendering for joint glow suppression and low-light enhancement. Cong et al.~\cite{cong2024semi,gui2026brightness} developed a spatial-frequency-aware semi-supervised framework, subsequently extended with brightness-aware synthetic-to-real learning. However, compared with hazy nighttime scenes, light effects in clear nighttime scenes are more spatially localized, posing greater challenges to accurate light-effect suppression and content preservation.

Different from the above unsupervised methods, our approach avoids explicit decomposition and prior-driven losses. Instead, it estimates a spatially varying light-effect map using a prior-embedded and dynamic integral-image estimator to guide the Transformer-based Schr\"odinger Bridge for light-effect suppression.

\vspace{-10pt}
\subsection{Schr\"odinger Bridge based methods}
In recent years, Schr\"odinger Bridge-based methods have shown promising performance in image-to-image translation. For example, I²SB~\cite{pmlr-v202-liu23ai} learns nonlinear diffusion bridges between two given domain distributions using paired data to perform image-to-image transfer. UNSB~\cite{kim2023unsb} proposes multi-step GAN models that solves the SB problem on unpaired data. Schr\"odinger Bridge Flow~\cite{de2024schrodinger} introduces an iterative flow of path measures to compute the SB, and demonstrates its effectiveness on unpaired image-to-image translation. CUNSB-RFIE~\cite{dong2025cunsb} constructs a context-aware Schr\"odinger Bridge framework using dynamic convolution for retinal fundus image enhancement. Lan et al.~\cite{lan2025schrodinger} achieved unpaired Schr\"odinger Bridge training for image dehazing by incorporating a CLIP-based discriminator.

Inspired by solving the Schr\"odinger Bridge with adversarial objectives~\cite{gushchin2023entropic,kim2023unsb}, we propose a cyclic Schr\"odinger Bridge framework for unpaired training, which achieves high-quality light-effect suppression under the guidance of an adaptive Gaussian-like prior.


\begin{figure*}[!htbp]
        \vspace{-18pt}
        \centering
        \includegraphics[width=1\linewidth]{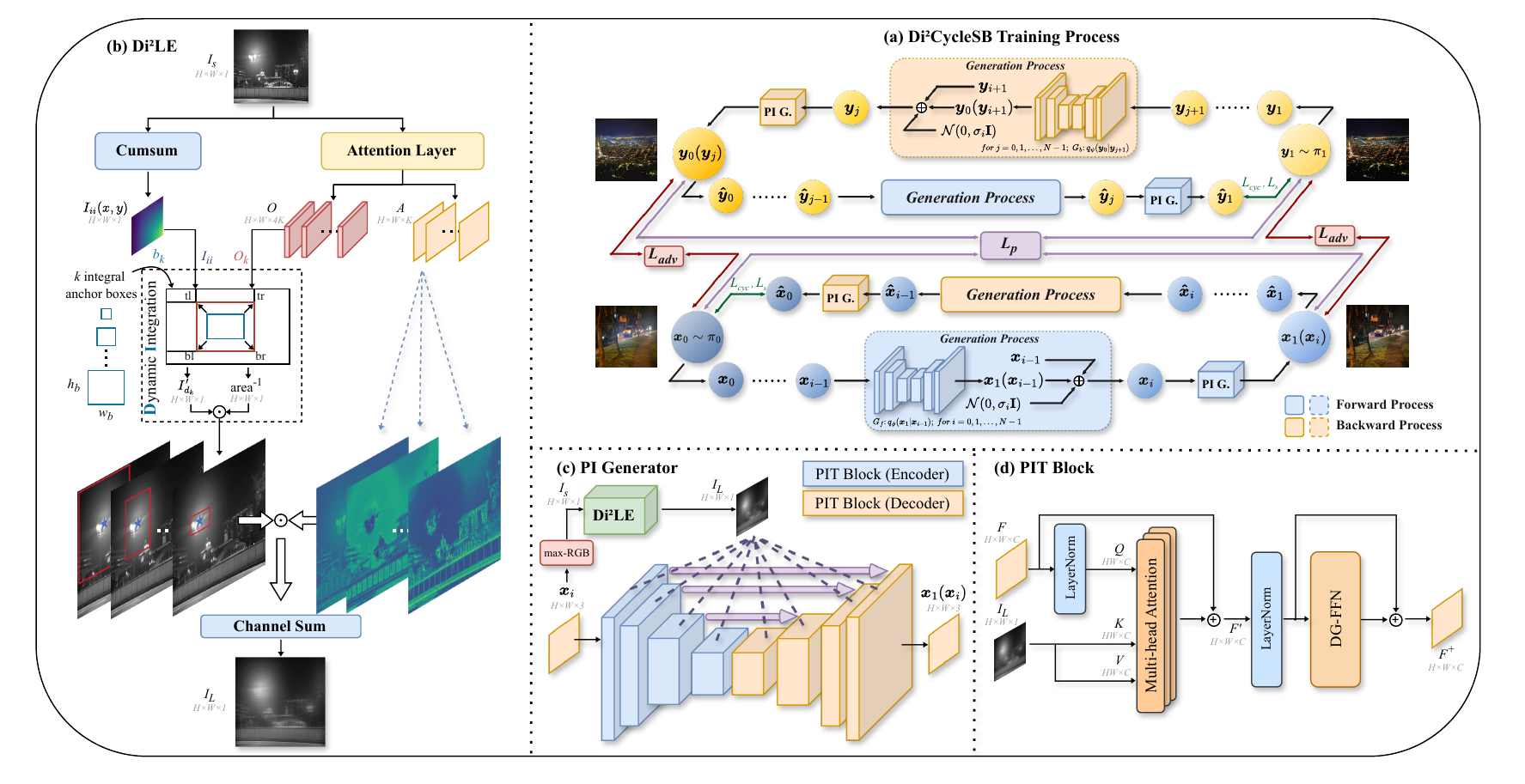}
        \vspace{-25pt}
        \caption{Overview of the Di²CycleSB framework. (a) illustrates the unpaired training process of Di²CycleSB. During training, we jointly optimize the forward and backward Schr\"odinger bridges at randomly sampled time steps and establish a cycle path to further regularize the transport process. (b) Di²LE estimates the light-effect map from the image intensity component. (c) The PI Generator performs light-effect suppression under the guidance of the estimated light-effect map. (d) The PIT Block is the core component of the PI Generator.}
        \label{fig:architecture}
        \vspace{-18pt}
\end{figure*}



\vspace{-15pt}
\section{Method}
\label{sec:method}
The overall architecture of our Dynamic-integral-image-Prior-guided Cycle Schr\"odinger Bridge Transformer (Di²CycleSB) framework is illustrated in Fig.~\ref{fig:architecture}. The framework consists of two stages where the first focuses on light-effect suppression and the second performs low-light image enhancement. Specifically, Fig.~\ref{fig:architecture} (a) illustrates the Cycle Schr\"odinger Bridge trained with unpaired data, where the generator is composed of Di²LE and PI Generator as depicted in Fig.~\ref{fig:architecture} (b) and (c) respectively. Fig.~\ref{fig:architecture} (d) further shows the details of the PIT Block, which is a key component of the PI Generator.

\vspace{-10pt}
\subsection{Di²CycleSB}
Since paired nighttime images are difficult to obtain, training with unpaired images has become a promising direction for light-effect suppression. However, substantial domain discrepancies make accurate distribution transport challenging. Unlike conventional unpaired generative models, the Schr\"odinger Bridge (SB) directly addresses entropy-regularized optimal transport (OT)~\cite{leonard2013survey} between two prescribed distributions. Least-cost transport provides a natural inductive bias toward localized degradation removal with minimal content alteration, facilitating high-quality light-effect suppression with few inference steps. Furthermore, inspired by~\cite{zhu2017unpaired}, we design forward and backward SBs to inject cycle consistency during training, further improving restoration quality under imperfect supervision. To this end, we formulate nighttime light-effect suppression as a Schr\"odinger Bridge problem (SBP), where the distributions of nighttime images with and without light effects are denoted by $\pi_0$ and $\pi_1$, respectively. Accordingly, we define the forward and backward stochastic processes as $\{\boldsymbol{x}_t\}$ and $\{\boldsymbol{y}_t\}$, with $t \in [0,1]$. Given a partition $\{t_i\}_{i=0}^{N}$ of the interval $[0,1]$, where $t_0=0 < t_{i-1} < t_i < t_N = 1$ and $N=5$ in this work, we shorthand  $\boldsymbol{x}_i := \boldsymbol{x}_{t_i}$, $\boldsymbol{x}_1 := \boldsymbol{x}_{t_N}$, $\boldsymbol{y}_i := \boldsymbol{y}_{t_i}$ and $\boldsymbol{y}_1 := \boldsymbol{y}_{t_N}$. According to the Markov chain decomposition, we can simulate the forward and backward SB as follows.
\begin{equation}
\begin{aligned}
p(\{\boldsymbol{x}_i\}_{i=0}^{N})
&= p(\boldsymbol{x}_{0})
\prod_{i=1}^{N} p(\boldsymbol{x}_{i} | \boldsymbol{x}_{i-1}), \\
p(\{\boldsymbol{y}_i\}_{i=0}^{N})
&= p(\boldsymbol{y}_{1})
\prod_{i=N}^{1} p(\boldsymbol{y}_{i-1} | \boldsymbol{y}_{i}).
\end{aligned}
\end{equation}

Given the known initial distributions $p(\boldsymbol{x}_{0})$ and $p(\boldsymbol{y}_{1})$, corresponding to night images with and without light effects, and assuming that samples from $p(\boldsymbol{x}_i)$ and $p(\boldsymbol{y}_i)$ are available, the above factorization enables us to recursively learn the transition probabilities $p(\boldsymbol{x}_i | \boldsymbol{x}_{i-1})$ and $p(\boldsymbol{y}_{i-1} | \boldsymbol{y}_i)$ to construct the forward and backward Markov processes. This procedure allows us to model forward-backward SB between $\pi_0$ and $\pi_1$.

For the forward SB, we define $q_{\phi_i}(\boldsymbol{x}_1 | \boldsymbol{x}_i)$ as a conditional distribution with learnable parameters $\phi_i$, with $q_{\phi_i}(\boldsymbol{x}_i, \boldsymbol{x}_1) := q_{\phi_i}(\boldsymbol{x}_1 | \boldsymbol{x}_i)\, p(\boldsymbol{x}_i)$ and $q_{\phi_i}(\boldsymbol{x}_1) := \mathbb{E}_{p(\boldsymbol{x}_i)}
\big[\, q_{\phi_i}(\boldsymbol{x}_1 | \boldsymbol{x}_i) \,\big]$. We specify the conditional transition as follow.
\begin{equation}
    \label{eq:sampling}
\begin{aligned}
p(\boldsymbol{x}_{{i+1}} | \boldsymbol{x}_1, \boldsymbol{x}_{i}) \sim \mathcal{N}\Big(\boldsymbol{x}_{{i+1}} ; s_{i+1} \boldsymbol{x}_1 + (1 - s_{i+1}) \boldsymbol{x}_{i},  \\ 
s_{i+1}(1 - s_{i+1})\, \tau (1 - t_i)\mathbf{I}\Big),
\end{aligned}
\end{equation}
where $s_{i+1} := (t_{i+1} - t_i)/(1 - t_i)$. As proved in \cite{kim2023unsb}, for any $t_i$, the optimal parameters $\phi_i$ can be obtained by solving the following constrained optimization problem.
\begin{equation}
\begin{aligned}
\min_{\phi_i}\; \mathcal{L}_{\mathrm{SB}}(\phi_i, t_i)
:= &\mathbb{E}_{q_{\phi_i}(\boldsymbol{x}_{i}, \boldsymbol{x}_1)} \left[\lVert \boldsymbol{x}_{i} - \boldsymbol{x}_1 \rVert_2^2\right]   \\
&- 2\tau (1 - t_i)\, \mathcal{H}\!\left(q_{\phi_i}(\boldsymbol{x}_{i}, \boldsymbol{x}_1)\right), \\
\end{aligned}
\end{equation}
\begin{equation}
\text{s.t.}\quad \mathcal{L}_{\mathrm{adv}}(\phi_i, t_i)
:= D_{\mathrm{KL}}\!\left(q_{\phi_i}(\boldsymbol{x}_1)\,\|\,p(\boldsymbol{x}_1)\right) = 0.
\end{equation}

Similarly, by defining $q_{\psi_i}(\boldsymbol{y}_0 | \boldsymbol{y}_i)$ with learnable parameters $\psi_i$, we can obtain the optimization objective of the backward SB:
\begin{equation}
\begin{aligned}
\min_{\psi_i}\; \mathcal{L}_{\mathrm{SB}}(\psi_i, t_i)
:= &\mathbb{E}_{q_{\psi_i}(\boldsymbol{y}_{i}, \boldsymbol{y}_0)} \left[\lVert \boldsymbol{y}_{i} - \boldsymbol{y}_0 \rVert_2^2\right]   \\
&- 2\tau t_i\, \mathcal{H}\!\left(q_{\psi_i}(\boldsymbol{y}_{i}, \boldsymbol{y}_0)\right), \\
\end{aligned}
\end{equation}
\begin{equation}
\text{s.t.}\quad \mathcal{L}_{\mathrm{adv}}(\psi_i, t_i)
:= D_{\mathrm{KL}}\!\left(q_{\psi_i}(\boldsymbol{y}_0)\,\|\,p(\boldsymbol{y}_0)\right) = 0.
\end{equation}

In practice, as shown in Fig.~\ref{fig:architecture}, we use time-conditional \hyperref[sec:PIG]{PI Generators} $G_f$ and $G_b$, whose parameters are shared across all time steps, to parameterize $q_{\phi}(\boldsymbol{x}_1 | \boldsymbol{x}_{i})$ and $q_{\psi}(\boldsymbol{y}_0 | \boldsymbol{y}_i)$, and iteratively perform the generation process by sampling the next state from the distribution according to Eq.~(\ref{eq:sampling}), thereby gradually suppressing light effects in night images.

\noindent \textbf{Training Process.}
Fig.~\ref{fig:architecture} (a) illustrates the training procedure of Di²CycleSB. Specifically, similar to the diffusion models, at each iteration, we randomly select a time step $t_i$ with $i\in{0,\dots,N-1}$ for training. Notably, the forward and backward time steps are independently sampled during training stage. This enables cycle consistency across intermediate states at different time steps, preventing degeneration into fixed stage-wise mappings. Moreover, varying $(t_i,t_j)$ exposes the generators to diverse input distributions, thereby promoting semantically meaningful bidirectional correspondences and mitigating the inherent steganographic effect~\cite{chu2017cyclegan}.

\noindent \textbf{Schr\"odinger Bridge Loss.} For each time step of the forward and backward processes, we transform the constrained optimization objective into a loss function using Lagrange multipliers:
$\mathcal{L} := \mathcal{L}_{\mathrm{adv}} + \lambda_{\mathrm{SB}} \mathcal{L}_{\mathrm{SB}}$.
Here, $\mathcal{L}_{\mathrm{adv}}$ is estimated via adversarial learning, while the entropy term in $\mathcal{L}_{\mathrm{SB}}$ is calculated via a mutual information estimator.

\noindent \textbf{Cycle Consistency Loss.}
Despite the inherent optimal transport properties of the Schr\"odinger Bridge, suppressing highly localized light effects using unpaired data remains challenging in terms of maintaining image content consistency, with potential failures manifested as artifacts. During training, we jointly optimize the forward and backward SBs under cycle consistency constraints. Specifically, we propagate the input images $\boldsymbol{x}_0\sim\pi_0$ and $\boldsymbol{y}_1\sim\pi_1$ along the forward and backward SBs, respectively, and sample their intermediate states $\boldsymbol{x}_i$ and $\boldsymbol{y}_i$ at an arbitrary time step $t_i$. They are then passed through $G_f$ and $G_b$ to obtain the target images $\boldsymbol{x}_1(\boldsymbol{x}_i)$ and $\boldsymbol{y}_0(\boldsymbol{y}_i)$, which are further required to reconstruct the input images at the same time step $t_i$. The cycle consistency loss is formulated as follows.
\begin{equation}
\begin{aligned}
    \mathcal{L}_\mathrm{cyc}(\phi, \psi, t_i) &:= \mathbb{E}_{\boldsymbol{x_0} \sim \pi_0, \boldsymbol{x_i} \sim p(\boldsymbol{x_i})}[\lVert \boldsymbol{\hat{x}_0}(\boldsymbol{x_i}) - \boldsymbol{x_0}\rVert_1]    \\
    &+ \mathbb{E}_{\boldsymbol{y_1} \sim \pi_1, \boldsymbol{y_i} \sim p(\boldsymbol{y_i})}[\lVert \boldsymbol{\hat{y}_1}(\boldsymbol{y_i}) - \boldsymbol{y_1}\rVert_1],
\end{aligned}
\end{equation}
where $\boldsymbol{\hat{x}_0}(\boldsymbol{x_i})=G_b(G_f(\boldsymbol{x_i}, t_i), t_i)$ and $\boldsymbol{\hat{y}_1}(\boldsymbol{y_i})=G_f(G_b(\boldsymbol{y_i}, t_i), t_i)$.

\noindent \textbf{Content Consistency Loss.}
We further introduce perceptual loss~\cite{johnson2016perceptual} and illumination consistency loss to preserve image content. Specifically, the perceptual loss $\mathcal{L}_{\mathrm{p}}$ constrains textural and structural consistency between the input images and the conditional outputs of $q_{\phi}(\boldsymbol{x}_1 | \boldsymbol{x}_i)$ and $q_{\psi}(\boldsymbol{y}_0 | \boldsymbol{y}_i)$, formulated as follows, where $f(\cdot)$ denotes the VGG16 network.
\begin{equation}
\begin{aligned}
    \mathcal{L}_\mathrm{p}(\phi, \psi, t_i) &:= \mathbb{E}_{\boldsymbol{x_0} \sim \pi_0, \boldsymbol{x_i} \sim p(\boldsymbol{x_i})}[\lVert f(G_f(\boldsymbol{x_i}, t_i)) - f(\boldsymbol{x_0})\rVert_2] \\
    &+ \mathbb{E}_{\boldsymbol{y_1} \sim \pi_1, \boldsymbol{y_i} \sim p(\boldsymbol{y_i})}[\lVert f(G_b(\boldsymbol{y_i}, t_i)) - f(\boldsymbol{y_1})\rVert_2].
\end{aligned}
\end{equation}

Inspired by~\cite{shi2024zero}, we leverage illumination smoothness to enforce consistency between the input and reconstructed images, complementing the pixel-level $\mathcal{L}_{\mathrm{cyc}}$ while reducing local intensity fluctuations. The $\mathcal{L}_{\mathrm{s}}$ is defined as follows:
\begin{equation}
\begin{aligned}
    \mathcal{L}_\mathrm{s}(\phi, \psi, t_i) &:= \mathbb{E}_{\boldsymbol{x_0} \sim \pi_0, \boldsymbol{x_i} \sim p(\boldsymbol{x_i})}[\lVert \mathcal{G}_{\sigma}(\boldsymbol{\hat{x}_0}(\boldsymbol{x_i})) - \mathcal{G}_{\sigma}(\boldsymbol{x_0}) \rVert_2] \\
    &+ \mathbb{E}_{\boldsymbol{y_1} \sim \pi_1, \boldsymbol{y_i} \sim p(\boldsymbol{y_i})}[\lVert \mathcal{G}_{\sigma}(\boldsymbol{\hat{y}_1}(\boldsymbol{y_i})) - \mathcal{G}_{\sigma}(\boldsymbol{y_1})\rVert_2],
\end{aligned}
\end{equation}
where $\mathcal{G}_{\sigma}(\cdot)$ denotes the Gaussian filtering with a window size of $21\times21$.

\noindent \textbf{Total loss.}
Our proposed loss functions are simple and general, and the overall optimization objective can be formulated as follows.
\begin{equation}
    \mathcal{L} = \mathcal{L}_{\mathrm{adv}} + \lambda_{\mathrm{SB}} \mathcal{L}_{\mathrm{SB}} + \lambda_{\mathrm{cyc}} \mathcal{L}_{\mathrm{cyc}} + \lambda_{\mathrm{p}} \mathcal{L}_{\mathrm{p}} + \lambda_{\mathrm{s}} \mathcal{L}_{\mathrm{s}}.
\end{equation}

In summary, we formulate unpaired light-effect suppression as an SBP. By discretizing the SB into a Markov chain, the complex distribution mapping is decomposed into a sequence of tractable transformations. Moreover, we construct cycle Schr\"odinger bridges and introducing a cycle-consistency loss to constrain the intermediate states of the forward and backward Markov chains, thereby achieving high-quality nighttime image enhancement.

\vspace{-10pt}
\subsection{Di²LE} 
In nighttime scenes, glow typically exhibits low-frequency characteristics and spatial non-uniformity due to anisotropic light scattering. Existing methods rely on fixed, hand-crafted regularization terms, limiting their ability to estimate non-uniform light effects in real-world scenes. Therefore, our motivation is to design a learnable and spatially adaptive light-effect estimator with inherently embedded glow priors to guide optimal transport learning for light-effect suppression within our unsupervised framework. Early studies established a connection between generalized Gaussian distributions and the atmospheric point spread function (APSF)~\cite{4408899}, while repeated application of averaging filters can approximate a Gaussian filter with a prescribed standard deviation~\cite{5692551}. Inspired by these observations, we propose a Dynamic-integral-image-based Light-Effects Estimator (Di²LE). Specifically, we adaptively aggregate a set of dynamic integral-image representations to parameterize Gaussian-like priors with spatially varying receptive fields and variances, thereby enabling effective estimation of non-uniform light effects.


Specifically, the Di²LE consists of four stages: 1) calculation of integral image; 2) pixel-wise prediction of offsets and weights for $K$ integral anchor boxes; 3) construction of $K$ dynamic integral images; and 4) weighted aggregation to obtain final representation. We first apply a max-RGB operation on the input image to get an intensity image $I_s \in \mathbb{R}^{H \times W \times 1}$. Then the integral image $I_{ii}$ of $I_s$ is calculated as follow.
\begin{equation}
I_{ii}(x, y) = \sum_{\substack{x' \leq x \\ y' \leq y}} I_s(x', y'),
\end{equation}
\noindent where $I_{ii}(x,y)$ represents the summed-area table of $I_s$, accumulating all pixel values within the region above and to the left of $(x,y)$. For each pixel, we predefine $K$ integral anchor boxes $\mathcal{B}=\{(w_{b_k},h_{b_k})\}_{k=1}^{K}$ with different window sizes as initial integration regions. In this work, we empirically set $K=5$ with window sizes $(w_{b_k}, h_{b_k}) \in \{5 \times 5,11 \times 11,21 \times 21,31 \times 31,43 \times 43\}$, where $k=1,\ldots, K$. This heuristic initialization of integration regions, analogous to weight initialization in convolutional neural networks, guides the module to learn low-frequency Gaussian priors during the early stages of training.

We employ a lightweight attention layer $\mathcal{F}_{\mathrm{att}}$, with its detailed architecture presented in the Appendix, to predict, for each pixel, the offsets of the $K$ integral anchor boxes in the top, left, bottom, and right directions, denoted as $\mathbf{O} \in \mathbb{R}^{H \times W \times 4K}$, together with $K$ weighting coefficients $\mathbf{A} \in \mathbb{R}^{H \times W \times K}$ satisfying the sum-to-one constraint (i.e., $\sum_{k=1}^{K} A_k(x,y)=1$). Based on the predefined anchor boxes and predicted offsets, the integration window coordinates $(\text{tl}, \text{tr}, \text{bl}, \text{br})$ of $k$-th dynamic integral image are calculated as follows.
\begin{equation}
\begin{aligned}
\text{tl}_{x_k} &= x_c - l_k \cdot \frac{w_{b_k}}{2} \cdot s_w, \ \  \text{tl}_{y_k} = y_c - t_k \cdot \frac{h_{b_k}}{2} \cdot s_h, \\
\text{tr}_{x_k} &= x_c + r_k \cdot \frac{w_{b_k}}{2} \cdot s_w, \ \  \text{tr}_{y_k} = y_c - t_k \cdot \frac{h_{b_k}}{2} \cdot s_h, \\
\text{bl}_{x_k} &= x_c - l_k \cdot \frac{w_{b_k}}{2} \cdot s_w, \ \  \text{bl}_{y_k} = y_c + b_k \cdot \frac{h_{b_k}}{2} \cdot s_h, \\
\text{br}_{x_k} &= x_c + r_k \cdot \frac{w_{b_k}}{2} \cdot s_w, \ \  \text{br}_{y_k} = y_c + b_k \cdot \frac{h_{b_k}}{2} \cdot s_h,
\end{aligned}
\label{eq:cal_coor}
\end{equation}
\noindent where $(x_c, y_c)$ denotes the coordinates of a pixel in the image grid with ${(0,0),(0,1),\ldots,(x_c,y_c),\ldots,(W,H)}$. For the $k$-th dynamic integral image, $(t_k, l_k, b_k, r_k)$ are the predicted offsets from $\mathbf{O}$. To align the resolutions between training and inference, scaling factors $s_w = w / W$ and $s_h = h / H$ are applied, where $(w, h)$ and $(W, H)$ denote the training and actual input image resolutions, respectively.

Then, the $k$-th dynamic integral image $I_{d_k}$ can be calculated as follows.
\begin{equation}
\begin{aligned}
I'_{d_k}(x,y) &= I_{ii}(\text{br}_k) + I_{ii}(\text{tl}_k) - I_{ii}(\text{tr}_k) - I_{ii}(\text{bl}_k), \\
\text{area}(x, y) &= (t_k + b_k) \cdot (l_k + r_k) \cdot h \times w/4, \\
I_{d_k}(x,y) &= \frac{I'_{d_k}(x,y)}{\text{area}(x, y)}. 
\label{eq: integral}
\end{aligned}
\end{equation}

We obtain the $K$ dynamic integral images $I_d \in \mathbb{R}^{H \times W \times K}$ in parallel as described above. These representations can be interpreted as a family of low-pass filtering operators with spatially varying receptive fields. Finally, we aggregate them with one dynamic aggregation operator as follows.
\begin{equation}
I_L = \sum_{k=1}^{K}{I_{d_k} \cdot \mathbf{A}_k},
\end{equation}
where $I_L \in \mathbb{R}^{H \times W \times 1}$ denotes the estimated light-effect map. This operator enables adaptive integration of dynamic integral-image representations, yielding a learnable glow-aware prior at each pixel.

Overall, the proposed Di2LE develops a dynamic integral image algorithm with spatially adaptive receptive fields. A learnable aggregation operator is introduced to parameterize Gaussian-like priors, enabling effective estimation of non-uniform light effects without hand-crafted regularization constraints. The pipeline can be described as Alg. \ref{alg:di2le}.

\begin{figure*}[!t]
        \vspace{-18pt}
        \centering
        \includegraphics[width=\textwidth]{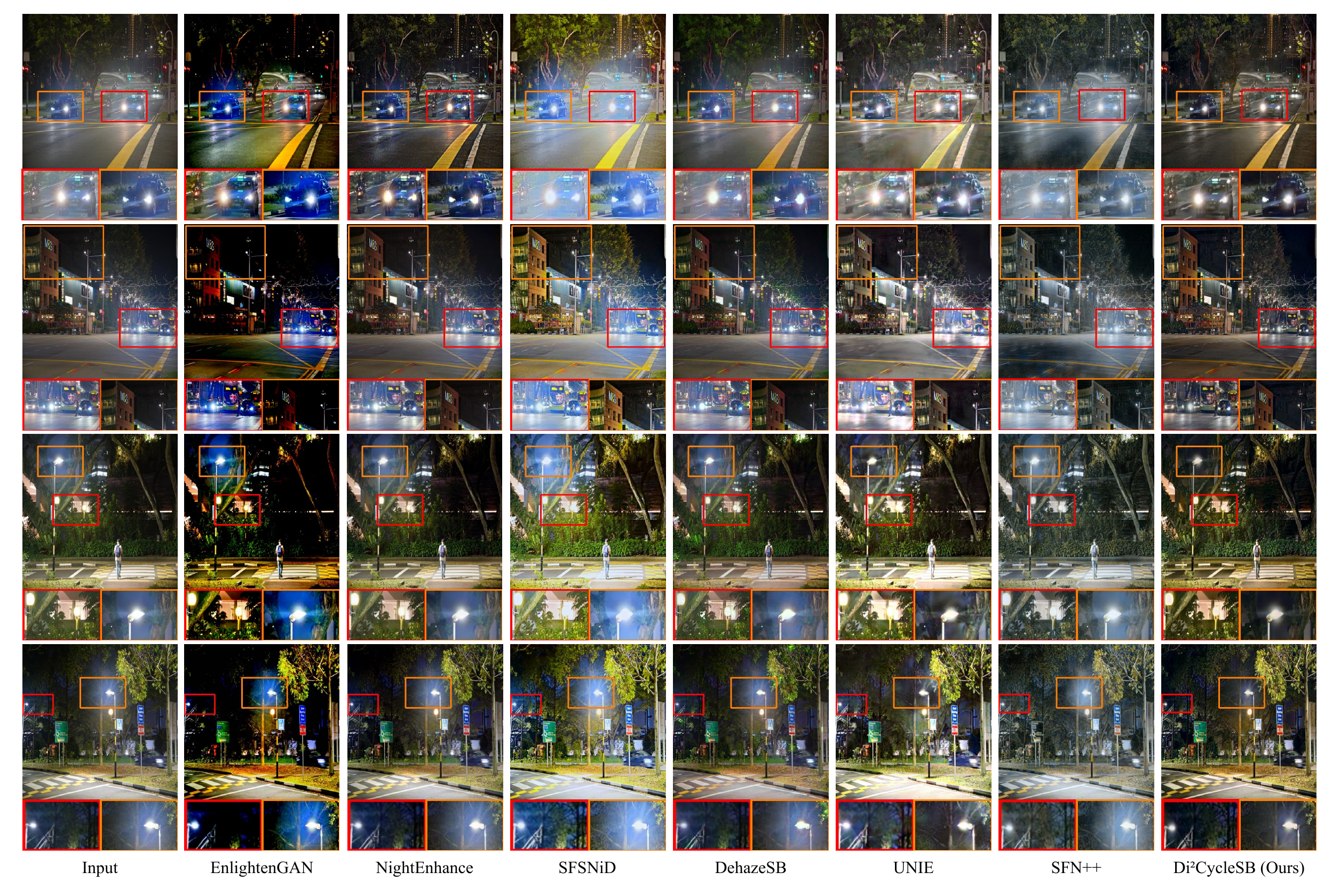}
        \vspace{-28pt}
        \caption{Visual qualitative comparisons of state-of-the-art methods on the Light-Effects dataset.}
        \label{fig:visual_results}
        \vspace{-13pt}
\end{figure*}

\begin{algorithm}[t]
\caption{Di²LE for light-effects Estimation}
\label{alg:di2le}
\renewcommand{\algorithmicrequire}{\textbf{Input:}}
\renewcommand{\algorithmicensure}{\textbf{Output:}}
\begin{algorithmic}[1]
    \REQUIRE{$I \in \mathbb{R}^{H \times W \times 3}$}
    \hfill $\triangleright$ Input Image
    \ENSURE{$I_L \in \mathbb{R}^{H \times W \times 1}$}
    \hfill $\triangleright$ Light-effects estimation of $I$
    \STATE Initialization: $\mathcal{B}=\{(w_{b_k},h_{b_k})\}_{k=1}^{K}$
    \STATE $I_s(x,y) = \max_{c \in \{R,G,B\}} I_c(x,y)$
    \STATE $I_{ii}(x,y) = \sum_{x' \leq x,\,y' \leq y} I_s(x',y')$
    \hfill $\triangleright$ Get integral image
    \STATE $(\mathbf{O},\mathbf{A}) = \mathcal{F}_{\mathrm{att}}(I_s)$
    \FOR{$k=1$ to $K$}
        \STATE $(t_k,l_k,b_k,r_k) = \mathbf{O}_k$
        \STATE Calculate $(\mathrm{tl}_k,\mathrm{tr}_k,\mathrm{bl}_k,\mathrm{br}_k)$ by Eq.~\eqref{eq:cal_coor}
        \STATE $I'_{d_k} =
        I_{ii}(\mathrm{br}_k)+I_{ii}(\mathrm{tl}_k)
        -I_{ii}(\mathrm{tr}_k)-I_{ii}(\mathrm{bl}_k)$
        \STATE $I_{d_k} = I'_{d_k}/\operatorname{area}_k$
    \hfill $\triangleright$ Get K dynamic integral images
    \ENDFOR
    \STATE $I_L = \sum_{k=1}^{K} \mathbf{A}_k \odot I_{d_k}$
    \hfill $\triangleright$ Dynamic aggregation
    \RETURN $I_L$
\end{algorithmic}
\end{algorithm}

\vspace{-12pt}
\subsection{Prior-Informed Generator}
\label{sec:PIG}
As illustrated in Fig.~\ref{fig:architecture} (c), PI Generator adopts a U-shaped architecture. Each stage is composed of a time-conditional Prior-Informed Transformer (PIT) block. In the encoder, the resolution of feature map is progressively reduced across four layers, i.e., $H \times W$, $\frac{H}{2} \times \frac{W}{2}$, $\frac{H}{4} \times \frac{W}{4}$, and $\frac{H}{8} \times \frac{W}{8}$. Correspondingly, in the decoder, the feature map resolution is gradually increased in a symmetric manner, i.e., $\frac{H}{8} \times \frac{W}{8}$, $\frac{H}{4} \times \frac{W}{4}$, $\frac{H}{2} \times \frac{W}{2}$, and $H \times W$. Skip connections are introduced to alleviate information loss caused by downsampling. an extra PIT block is placed after the decoder for refinement. The light-effect map $I_L$, estimated by the Di²LE module, is adapted to different feature resolutions using $4\times4$ convolutions with a stride of 2 and injected into each layer to guide light-effect suppression.

\begin{figure*}[!t]
        \vspace{-18pt}
        \centering
        \includegraphics[width=\textwidth]{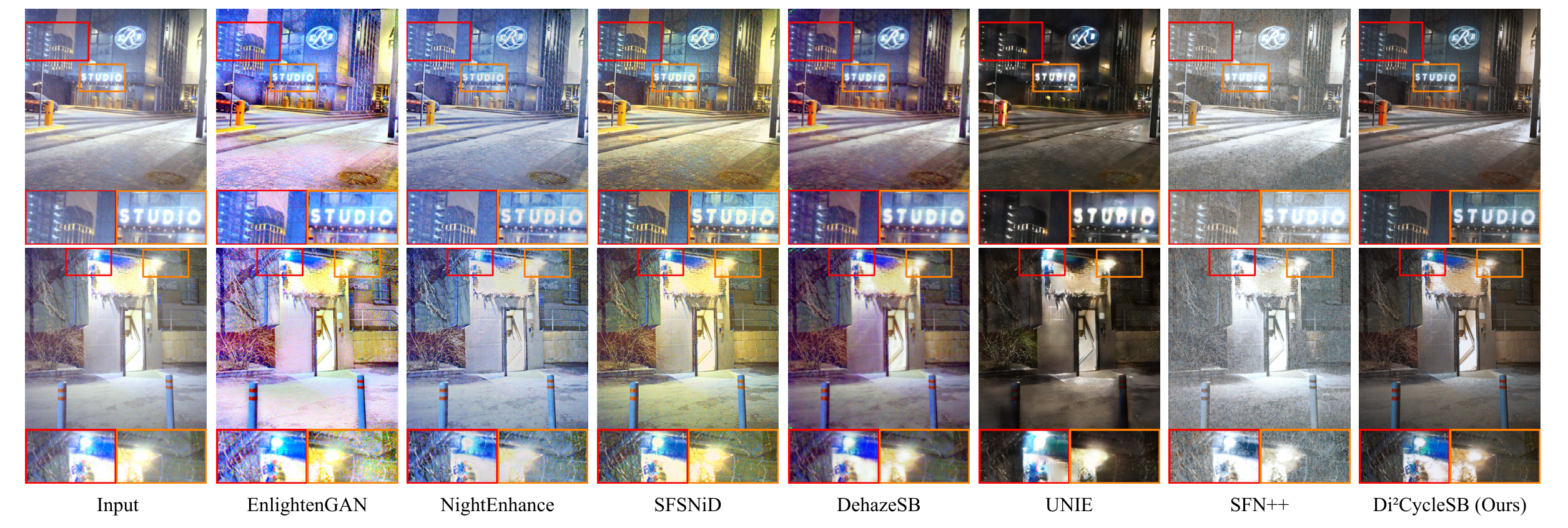}
        \vspace{-26pt}
        \caption{Visual qualitative comparisons of state-of-the-art methods on the NTIRE 2025 Challenge 2K resolution dataset.}
        \label{fig:visual_results_NIEU2k}
        \vspace{-20pt}
\end{figure*}

\noindent \textbf{PIT. } Each PIT Block comprises three time-conditional RMSNorm~\cite{zhang2019root} layers, one Multi-Head Cross Attention layer, and one Dual-Gated Feed-Forward Network (DG-FFN)~\cite{wang2023ultra} layer. Specifically, inspired by the success of cross-attention mechanisms in image editing and style transfer~\cite{deng2022stytr2}, we design a simple yet effective cross-attention module to integrate the knowledge in the estimated light-effect map into the feature representations. The input feature $F$ is treated as the query (Q), while the light-effect map $I_L$ is regarded as the key (K) and value (V):
\begin{equation}
\begin{aligned}
    \mathbf{Q} &= W_d W_p \, \mathrm{LN}(F, t_i),\\
    \mathbf{K}, \mathbf{V} &= \mathrm{Split}(W_d W_p \mathrm{LN}(I_L, t_i)),
\end{aligned}
\end{equation}
where $W_d$ and $W_p$ denote a $3 \times 3$ depthwise separable convolution and a $1 \times 1$ pointwise convolution, where $\operatorname{LN}(\cdot,t_i)$ denotes the time-conditional RMSNorm, in which the time-dependent scaling and bias parameters $\boldsymbol{\gamma}(t_i)$ and $\boldsymbol{\beta}(t_i)$ are predicted from the time step $t_i$ through two linear projections, respectively, and $\odot$ denotes element-wise multiplication.

\begin{equation}
\begin{aligned}
\boldsymbol{\gamma}(t_i)
&=
\operatorname{Sigmoid}\!\left(\operatorname{Linear}_{\gamma}(t_i)\right),\\
\boldsymbol{\beta}(t_i)
&=
\operatorname{Linear}_{\beta}(t_i),\\
\operatorname{LN}(\boldsymbol{x},t_i)
&=
\boldsymbol{\gamma}(t_i)\odot
\operatorname{RMSNorm}(\boldsymbol{x})
+
\boldsymbol{\beta}(t_i).
\end{aligned}
\label{eq:time_conditional_rmsnorm}
\end{equation}

By treating each single-channel feature map as a token, we compute channel-wise affinities to balance long-range dependency modeling with computational efficiency. Subsequently, channel-wise multi-head cross-attention is applied between $F$ and $I_L$, enabling the feature map $F$ to query informative light-effect cues from $I_L$ and update the query as follows.
\begin{equation}
\begin{aligned}
\text{Attention}_n(\mathbf{Q}, \mathbf{K}, \mathbf{V}) &= \mathbf{V} \cdot \mathrm{softmax}\!\left(\frac{\mathbf{K}^\top \mathbf{Q}}{\alpha}\right),                \\
\mathcal{F}_{\mathrm{MSA}}(\mathbf{Q}, \mathbf{K}, \mathbf{V}) &= W_p \, \mathrm{Concat}(\text{Attention}_1(\mathbf{Q}, \mathbf{K}, \mathbf{V}),  \\
&\qquad \ldots, \text{Attention}_N(\mathbf{Q}, \mathbf{K}, \mathbf{V})), \\
F' &= \mathcal{F}_{\mathrm{MSA}}(\mathbf{Q}, \mathbf{K}, \mathbf{V}) + \mathbf{Q},
\label{eq:attn}
\end{aligned}
\end{equation}
where $n = 1, 2, \ldots, N$ denotes the number of attention heads, and $\alpha$ is a learnable scaling factor. Notably, we adopt a transposed attention strategy that treats feature channels as tokens, in order to reduce the computational overhead of vanilla attention. Finally, the updated feature $F'$ integrates prior knowledge from the light-effect map and is further refined by the DG-FFN layer to suppress light effects:
\begin{equation}
\begin{aligned}
\mathcal{F}_{\text{DG-FFN}}(F') &=\ W_p(\text{GELU}(W_{p1} F') \odot W_{p2} F' \\
 &\qquad + \text{Sigmoid}(W_{p2} F') \odot W_{p1} F'),  \\
F^+ &= \mathcal{F}_{\text{DG-FFN}}(F') + F',
\end{aligned}
\end{equation}
where $W_p, W_{p1}$, and $W_{p2}$ denote $1 \times 1$ convolution layers.

Overall, the proposed PI Generator employs a carefully designed cross-attention mechanism that allows the backbone features to interact with the light-effect map in a multi-scale manner, thereby capturing non-local information associated with light effects and enabling effective light-effect suppression.

\vspace{-10pt}
\subsection{Low-Light Image Enhancement}
\vspace{-4pt}
Inspired by Zero-DCE~\cite{li2021learning}, we introduce an iterative light-effect-guided LLIE module to enhance illumination after light-effect suppression. Since regions affected by strong light effects are typically well illuminated and require minimal enhancement, we leverage the estimated light-effect map $I_L$ from the first stage to guide spatially adaptive enhancement. Specifically, given a light-effect-free image $I_{\mathrm{lf}} \in \mathbb{R}^{H \times W \times 3}$ from the first stage, a simple CNN is first employed to predict a pixel-wise parameter map $R \in \mathbb{R}^{H \times W \times 3}$. We then invert the light-effect map $I_L$ and apply it via element-wise multiplication to a higher-order image-specific curve, upon which low-light image enhancement is performed iteratively. The update rule is given as follow.
\begin{equation}
\begin{aligned}
\mathrm{LLIE}_m(x) &= \mathrm{LLIE}_{m-1}(x) + \\
&(1 - I_L(x))\, R(x)\, \mathrm{LLIE}_{m-1}(x)\big(1 - \mathrm{LLIE}_{m-1}(x)\big),
\end{aligned}
\end{equation}
where $m = 1, \ldots, M$  and $M$ is set to $3$ in this paper. $\mathrm{LLIE}_0(x)$ is defined as the identity mapping. And the CNN architecture and loss functions follow~\cite{li2021learning}, with details given in the Appendix.

\vspace{-10pt}
\section{Experiments and Results}
\label{sec:experiment}
\subsection{Datasets and Implementation Details}
\subsubsection{Datasets}
We evaluate our method on two representative datasets, i.e., Light-Effects dataset~\cite{sharma2021nighttime} and NTIRE 2025 Challenge dataset~\cite{Ershov_2025_CVPR}.

\noindent \textbf{Light-Effects. }
It is a standard unpaired nighttime image dataset, containing 501 real-world night images, which include diverse light effects with varying numbers, shapes, and colors.

\noindent \textbf{NTIRE 2025 Challenge. }
The dataset contains 1,000 paired $2\text{k}$-resolution nighttime RAW images with and without light effects. We convert the RAW images into RGB images using the processing pipeline in~\cite{Ershov_2025_CVPR}. We then select 500 light-effect-affected and 500 non-overlapping light-effect-free images to construct an unpaired high-resolution benchmark, which is released alongside our code.


\subsubsection{Implementation Details}
We jointly train the forward and backward SBs with $N=5$ time steps, using identical generator architectures $G_f$ and $G_b$ (Sec.~\ref{sec:PIG}) and Markovian discriminators for $\mathcal{L}_{\mathrm{Adv}}$. Di²CycleSB is trained for 100 epochs using Adam ($\beta_1=0.9$, $\beta_2=0.999$), with an initial learning rate of $2\times10^{-4}$ linearly decayed to $1\times10^{-6}$ over the final 50 epochs. The loss weights are fixed at $\lambda_{\mathrm{SB}}=1.0$, $\lambda_{\mathrm{cyc}}=5.0$, $\lambda_{\mathrm{p}}=0.5$, and $\lambda_{\mathrm{s}}=1.0$. In addition, we apply random cropping, flipping, and rotation. All experiments using $256\times 256$ training crops are performed on a single NVIDIA 4090 GPU.

After training, we adopt single-step forward SB inference for efficiency and use its outputs to train the LLIE module following~\cite{li2021learning}. During inference, the forward SB and LLIE module are cascaded for nighttime visibility enhancement.

\begin{table*}[!t]
    \renewcommand{\arraystretch}{1.3}
    \small 
    \vspace{-18pt}
    \centering
    \caption{Quantitative comparison on the \textbf{Light-Effects} dataset via user studies and no-reference IQA metrics.}
    \vspace{-8pt}
    \begin{tabular}{p{2.4cm}<{\centering} | p{1.3cm}<{\centering} | 
                    p{1.8cm}<{\centering} | p{3.cm}<{\centering} | 
                    p{1.6cm}<{\centering} | p{1.8cm}<{\centering} | 
                    p{1.5cm}<{\centering}| p{1.cm}<{\centering}}
        \Xhline{1pt}
        \multirow{2}{*}{\textbf{Methods}}
        & \multirow{2}{*}{\textbf{Venue}}
        & \multirow{2}{*}{\textbf{User Study$\uparrow$}}
        & \textbf{LMM-IQA$\uparrow$}
        & \multirow{2}{*}{\textbf{Q-Align$\uparrow$}}
        & \multirow{2}{*}{\textbf{CLIP-IQA+$\uparrow$}}
        & \multirow{2}{*}{\textbf{MANIQA$\uparrow$}}
        & \multirow{2}{*}{\textbf{LIQE$\uparrow$}}
        \\
        & & &
        \textbf{(sampled/full dataset)}
        & & & &
        \\
        \Xhline{1pt}
        Original Dataset & -- & -- & -- & 3.368 & 0.603 & 0.368 & 2.745 \\ 
        \Xhline{1pt}
        EnlightenGAN~\cite{jiang2021enlightengan} & TIP'21 
        & -1.158 & -1.635/-1.853 & 2.552 & 0.451 & 0.330 & 1.991                                      \\ \hline
        Dehazeformer~\cite{song2023vision} & TIP'23 
        & -1.491 & -1.011/-0.849 & 3.015 & 0.513 & 0.321 & 2.399                                        \\ \hline
        UNIE~\cite{jin2022unsupervised} & ECCV'22 
        & 0.356 & -0.164/-1.991 & 3.070 & 0.480 & 0.274 & 1.839                                         \\ \hline
        
        NightEnhance~\cite{jin2023enhancing} & MM'23 
        & \underline{0.855} & \textit{0.770}/\textit{0.763} & \underline{3.469} & 0.567 & \textit{0.334} & \underline{2.574}                                         \\ \hline
        SFSNiD~\cite{cong2024semi} & CVPR'24 
        & -0.955 & -0.603/-0.453 & 2.970 & \textit{0.574} & 0.306 & 2.329                                        \\ \hline

        DehazeSB~\cite{lan2025schrodinger}  & ICCV'25 
        & \textit{0.702} & 0.224/0.517 & \textit{3.269} & 0.540 & 0.322 & \textit{2.412}                                                                             \\  \hline

        SFN++~\cite{gui2026brightness}  & TPAMI'26 
        & 0.233 & \underline{0.790}/\underline{0.898} & 3.172 & \underline{0.578} & \underline{0.357} & 2.147                                                                            \\  \hline

        \textbf{Di²CycleSB} & --
        & \textbf{1.457} & \textbf{1.627}/\textbf{1.204} & \textbf{3.567} & \textbf{0.644} & \textbf{0.360} & \textbf{2.929}                                                                             \\
        \Xhline{1pt}
    \end{tabular}
    \label{tab:lighteffects-result}
\end{table*}

\begin{table*}[!t]
    \renewcommand{\arraystretch}{1.3}
    \small 
    \centering
    \vspace{-10pt}
    \caption{Quantitative comparison on the \textbf{NTIRE 2025 Challenge} unpaired dataset via user studies and no-reference IQA metrics.}
    \vspace{-8pt}
    \begin{tabular}{p{2.4cm}<{\centering} | p{1.3cm}<{\centering} | 
                    p{1.8cm}<{\centering} | p{3.cm}<{\centering} | 
                    p{1.6cm}<{\centering} | p{1.8cm}<{\centering} | 
                    p{1.5cm}<{\centering}| p{1.cm}<{\centering}}
        \Xhline{1pt}
        \multirow{2}{*}{\textbf{Methods}}
        & \multirow{2}{*}{\textbf{Venue}}
        & \multirow{2}{*}{\textbf{User Study$\uparrow$}}
        & \textbf{LMM-IQA$\uparrow$}
        & \multirow{2}{*}{\textbf{Q-Align$\uparrow$}}
        & \multirow{2}{*}{\textbf{CLIP-IQA+$\uparrow$}}
        & \multirow{2}{*}{\textbf{MANIQA$\uparrow$}}
        & \multirow{2}{*}{\textbf{LIQE$\uparrow$}}
        \\
        & & &
        \textbf{(sampled/full dataset)}
        & & & &
        \\
        \Xhline{1pt}

        Original Dataset & -- & -- & -- & 2.241 & 0.293 & 0.175 & 1.186 \\ \hline
        \Xhline{1pt}

        EnlightenGAN~\cite{jiang2021enlightengan} & TIP'21 
        & -0.678 & -1.640/-1.700 & 1.603 & 0.301 & 0.156 & 1.164           \\ \hline

        Dehazeformer~\cite{song2023vision} & TIP'23 
        & -0.416 & -0.695/-0.667 & 1.960 & 0.306 & 0.153 & 1.154          \\ \hline

        UNIE~\cite{jin2022unsupervised}    & ECCV'22 
        & \underline{0.845}  & \underline{1.014}/\underline{0.878} & \underline{2.609} & \textit{0.315} & \underline{0.188} & \underline{1.706}          \\ \hline
        
        NightEnhance~\cite{jin2023enhancing} & MM'23 
        & 0.206 & 0.222/0.093 & 1.942 & \underline{0.325} & 0.152 & 1.168          \\ \hline

        SFSNiD~\cite{cong2024semi}  & CVPR'24 
        & 0.135 & -0.208/-0.120 & \textit{2.204} & 0.283 & 0.164 & \textit{1.353}          \\ \hline

        DehazeSB~\cite{lan2025schrodinger}  & ICCV'25 
        & \textit{0.813} & \textit{0.421}/\textit{0.437} & 2.110 & 0.284 & 0.134 & 1.124           \\ \hline

        SFN++~\cite{gui2026brightness}  & TPAMI'26 
        & -2.114 & -0.810/-0.706 & 1.931 & 0.357 & \textit{0.179} & 1.163                                                                            \\  \hline

        \textbf{Di²CycleSB} & --     
        & \textbf{1.210} & \textbf{1.696}/\textbf{1.784} & \textbf{2.761} & \textbf{0.412} & \textbf{0.198} & \textbf{2.066}             \\

        \Xhline{1pt}
    \end{tabular}
    \label{tab:ntire-result}
    \vspace{-15pt}
\end{table*}

\vspace{-10pt}
\subsection{Comparisons with State-of-the-Art Methods}
To validate the effectiveness of Di²CycleSB, we conduct comprehensive comparisons using User Study, Proprietary Large Multimodal Models (LMMs) Evaluation, and Image Quality Assessment (IQA). For a fair comparison, all unsupervised and semi-supervised methods are retrained using their officially recommended settings. In all experimental tables, the best, second-best, and third-best results are indicated by \textbf{boldface}, \underline{underlining}, and \textit{italics}, respectively.

\noindent \textbf{Visual Results. }
Fig.~\ref{fig:visual_results} and Fig.~\ref{fig:visual_results_NIEU2k} present visual comparisons (zoom in for better viewing). Low-light image enhancement method EnlightenGAN~\cite{jiang2021enlightengan} overlooks light-effect suppression, resulting in color distortions and residual light effects. Semi-supervised nighttime dehazing methods (NightEnhance~\cite{jin2023enhancing}, SFSNiD~\cite{cong2024semi}, and SFN++~\cite{gui2026brightness}) suffer from inaccurate pseudo-labels and domain gaps, leading to insufficient light-effect suppression or color distortions. DehazeSB~\cite{lan2025schrodinger} struggles with spatially non-uniform light effects and introduces unnatural color alterations, including overly green trees and global color shifts. UNIE~\cite{jin2022unsupervised} produces artifacts due to its reliance on hand-crafted priors and image decomposition. In contrast, Di²CycleSB effectively suppresses light effects while preserving natural colors and enhancing visibility, even in high-resolution images.

\noindent \textbf{User Study. }
\label{sec:user_study}
In the absence of widely accepted objective metrics for light-effect suppression, we conduct comprehensive user studies on Light-Effects and NTIRE 2025 Challenge datasets, respectively. For each dataset, we randomly select 50 images and invite twelve participants to rank eight methods based on light-effect suppression, visual naturalness and consistency, and visibility enhancement. Further implementation details are provided in the Appendix. Following ranking-based NR-IQA methods~\cite{9548827}, the rankings are converted into pairwise preferences and aggregated using maximum a posteriori estimation under the Thurstone Case V model~\cite{thurstone1927law}. As shown in Tab.~\ref{tab:lighteffects-result} and Tab.~\ref{tab:ntire-result}, Di²CycleSB achieves the highest user study scores on both datasets.

\noindent \textbf{Proprietary Large Multimodal Models Evaluation.}
\label{sec:LMM}
Although human studies provide reliable image quality assessments, evaluating large-scale datasets is prohibitively labor-intensive. Given the human-level performance of recent proprietary LMMs on complex visual tasks~\cite{feng2026seeing}, we employ the state-of-the-art GPT-5.6 Sol~\cite{openai2025gpt5} for LMM-based image quality assessment (LMM-IQA). Following~\cite{feng2026seeing}, we formulate the human user study criteria as a globally consistent system prompt and instruct GPT-5.6 Sol to rank the enhancement results of all methods for each image. The rankings are then converted into scores using the Thurstone Case V model, with implementation details provided in the Appendix. In Tab.~\ref{tab:lighteffects-result} and Tab.~\ref{tab:ntire-result}, scores before and after "\texttt{/}" correspond to the 50 images sampled for the human study and the full dataset, respectively. The results demonstrate strong agreement with human judgments, while our method achieves the highest scores on both full datasets, substantially outperforming the second-best method.

\noindent \textbf{Image Quality Assessment. }
Furthermore, we adopt four widely used IQA metrics, including Q-Align~\cite{wu2024qalign}, CLIP-IQA+~\cite{wang2023clipiqa}, MANIQA~\cite{yang2022maniqa}, and LIQE~\cite{zhang2023liqe}. As shown in Tab.~\ref{tab:lighteffects-result} and Tab.~\ref{tab:ntire-result}, Di²CycleSB consistently achieves the best performance across all IQA metrics on both datasets, with particularly notable improvements in LIQE on the NTIRE 2025 Challenge dataset. These results demonstrate its effectiveness and generalizability in suppressing light effects while preserving perceptual image quality.

\begin{table*}[!htbp]
\centering
\renewcommand{\arraystretch}{1.15}
\vspace{-18pt}
\caption{Breakdown ablation study of the proposed framework.}
\vspace{-8pt}
\label{tab:breakdown_ablation}

\begin{tabular}{
p{3.2cm}<{\centering}|
p{2.0cm}<{\centering}
p{2.0cm}<{\centering}
p{2.0cm}<{\centering}
p{2.0cm}<{\centering}
p{2.0cm}<{\centering}}
\hline
\textbf{Experiments}
& \textbf{GPT-5.6 Sol$\uparrow$}
& \textbf{Q-Align$\uparrow$}
& \textbf{CLIP-IQA+$\uparrow$}
& \textbf{MANIQA$\uparrow$}
& \textbf{LIQE$\uparrow$} \\
\hline

w/o Cycle
& \textit{0.556} & \textit{3.516} & 0.581 & 0.338 & \textit{2.854} \\

w/o SB
& -1.157 & 3.391 & 0.561 & 0.342 & 2.719 \\

w/o Di²LE
& 0.467 & 3.463 & \textit{0.629} & \textit{0.341} & 2.752 \\

w/o PI generator
& -1.601 & 2.722 & 0.478 & 0.190 & 1.373 \\

w/o LLIE
& \underline{0.681} & \underline{3.547} & \underline{0.634} & \underline{0.348} & \underline{2.899} \\

\hline
\textbf{Di²CycleSB}
& \textbf{1.053}
& \textbf{3.567}
& \textbf{0.644}
& \textbf{0.360}
& \textbf{2.929} \\
\hline
\end{tabular}
\vspace{-15pt}
\end{table*}

\begin{figure}[!htbp]
        \centering
        \vspace{-10pt}
        \includegraphics[width=\linewidth]{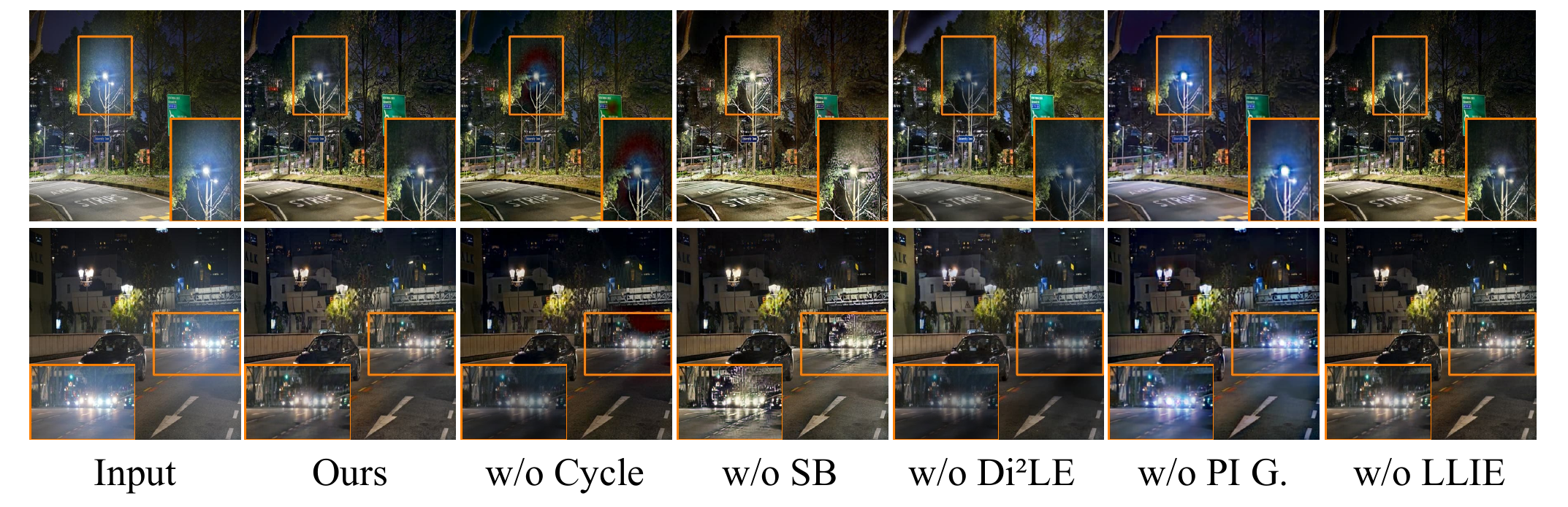}
        \vspace{-25pt}
        \caption{Qualitative visual results of the ablation study. Please zoom in for better viewing.}
        \label{fig:ablation}
        \vspace{-5pt}
\end{figure}

\vspace{-18pt}
\subsection{Ablation Study}
To demonstrate the effectiveness of each module in the proposed Di²CycleSB framework, we conduct extensive ablation studies on the Light-Effects dataset~\cite{sharma2021nighttime}.

\noindent \textbf{Break-down Ablation. }
To validate the effectiveness of our key contributions, we conduct five ablation studies. 
\begin{itemize}
    \item ``w/o Cycle'' removes the backward SB and consistency losses $\mathcal{L}_{\mathrm{cyc}}$ and $\mathcal{L}_{\mathrm{s}}$;
    \item ``w/o SB'' removes the SB formulation from both processes while retaining the losses and backbone;
    \item ``w/o Di²LE'' removes the Di²LE module;
    \item ``w/o PI Generator'' replaces the PI Generator with a 9-layer time-conditional ResNet using concatenated image and light-effect map $I_L$ as input;
    \item ``w/o LLIE'' removes the final $I_L$-guided enhancement stage.
\end{itemize}

As shown in Tab.~\ref{tab:breakdown_ablation}, removing any key component degrades performance across all evaluation metrics. Notably, ``w/o LLIE'' achieves the second-best results, highlighting the effectiveness of the core Di²CycleSB framework. The visual comparisons in Fig.~\ref{fig:ablation} further reveal that ``w/o Cycle'' compromises content consistency despite effective light-effect suppression, while ``w/o SB'' introduces color distortions due to the single-stage GAN architecture. Without the spatially adaptive Gaussian-like prior, ``w/o Di²LE'' fails to accurately suppress light effects, whereas ``w/o PI G.'' struggles to restore degraded regions due to limited long-range modeling capability. Finally, ``w/o LLIE'' confirms the necessity of the LLIE stage in compensating for the limited illumination enhancement capability of the training data.

\begin{table}[!htbp]
\centering
\vspace{-15pt}
\caption{Ablation study of the Di²LE module.}
\vspace{-6pt}
\label{tab:dile_ablation}
\renewcommand{\arraystretch}{1.18}
\setlength{\tabcolsep}{10pt}

\begin{tabular}{c c}
\hline
\textbf{Experiments} & \textbf{GPT-5.6 Sol$\uparrow$} \\
\hline
Max-RGB input                            & -0.295 \\
Gaussian kernel ($5\times5$)             & -0.372 \\
Gaussian kernel ($11\times11$)           & \underline{0.795} \\
Gaussian kernel ($21\times21$)           & \textit{0.184} \\
Gaussian kernel ($31\times31$)           & -0.011 \\
Gaussian kernel ($43\times43$)           & -1.892 \\
\hline
\textbf{Di²LE}                        & \textbf{1.591} \\
\hline
\end{tabular}
\vspace{-8pt}
\end{table}

\noindent \textbf{Di²LE. }
We consider a series of alternative designs to evaluate the effectiveness of the key components in the proposed Di²LE module. Specifically, we replace the proposed Di²LE module with six alternative light-effect estimation schemes, including five Gaussian filters with fixed kernel sizes of $5\times 5, 11\times 11, 21\times 21, 31\times 31$, and $43\times 43$, as well as the original Max-RGB input. The fixed kernel sizes are identical to the initial anchor-window sizes adopted by Di²LE. We then rank these variants using the proprietary LMM-based evaluation protocol. As reported in Tab.~\ref{tab:dile_ablation}, the proposed Di²LE achieves the highest score among all variants, demonstrating that its spatially adaptive Gaussian-like prior is the key to effective light-effect suppression. Importantly, as illustrated in the second row of Fig.~\ref{fig:ablation}, none of the variants can completely suppress light effects, whereas our method produces reasonable enhancement results.

\begin{figure}[!htbp]
    \centering
    \vspace{-8pt}
    \includegraphics[width=\linewidth]{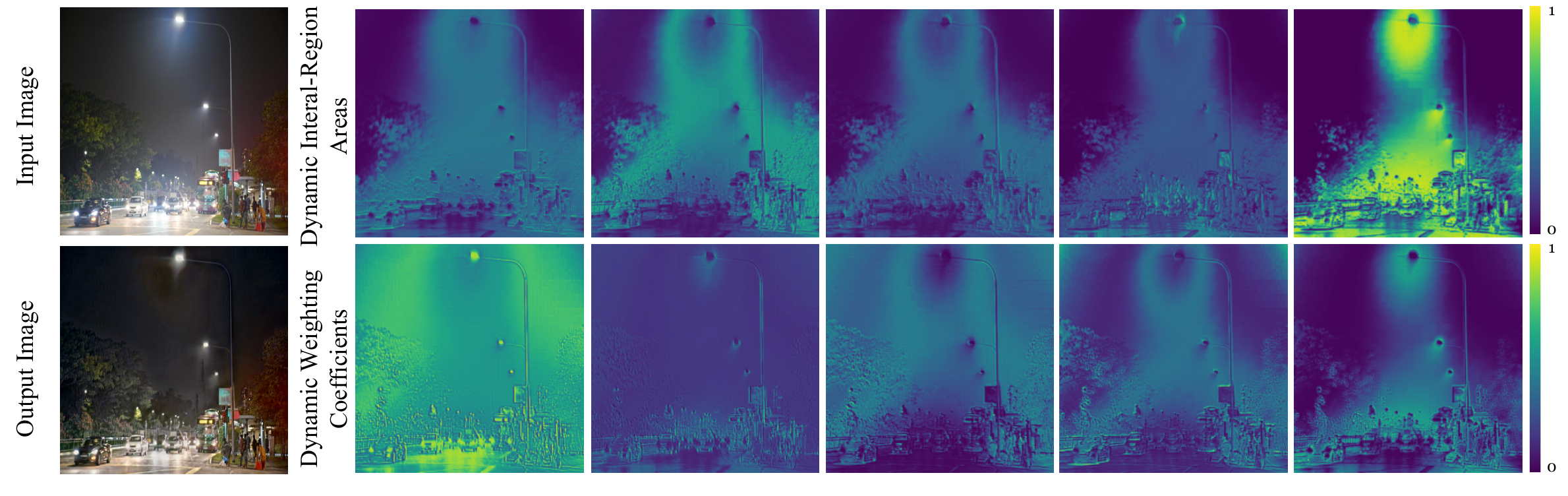}
    \vspace{-24pt}
    \caption{Normalized heatmap visualizations of the $K=5$ dynamic integral regions (top row) and the corresponding dynamic weighting coefficients (bottom row). The top row shows that pixels closer to light-effect regions activate larger low-pass filtering windows, whereas pixels farther away from light effects tend to activate smaller ones.}
    \label{fig:di2le}
\end{figure}


\vspace{-25pt}
\subsection{Visualization Analysis}
As described above, Di²LE employs dynamic integral images to emulate spatially varying low-pass filters and adaptively aggregates their representations to estimate Gaussian-like priors with varying receptive fields and variances. As shown in the first row of Fig.~\ref{fig:di2le}, the $K$ dynamic integral-image representations capture the non-uniform distribution of light effects, assigning larger integral regions to heavily affected areas and smaller ones to less affected regions. The second row further visualizes the adaptive weighting coefficients, demonstrating the spatially adaptive aggregation of integral-image representations for pixel-wise light-effect estimation.

\vspace{-10pt}
\section{Conclusion}
\label{sec:conclusion}
In this work, we propose Di²CycleSB, a novel unsupervised framework for suppressing light effects and enhancing nighttime visibility. We first introduce Di²LE, a light-effect estimator that parameterizes spatially adaptive Gaussian-like priors for glow estimation without relying on hand-crafted constraints. We further propose the PI Generator for end-to-end light-effect suppression under the guidance of the estimated light-effect priors. Building upon these components, we formulate light-effect suppression as a Schr\"odinger bridge problem and jointly optimize the forward and backward transport processes to introduce cycle consistency, achieving high-quality light-effect suppression. Extensive experiments on unpaired datasets demonstrate the superiority of the proposed method.

\vspace{-15pt}


\clearpage
\vspace*{-38pt}
\renewcommand{\IEEEbibitemsep}{-0.4pt}
\bibliographystyle{IEEEtran}
\bibliography{references}

@ARTICLE{10541091,
  author={Dai, Yuekun and Li, Chongyi and Zhou, Shangchen and Feng, Ruicheng and Luo, Yihang and Loy, Chen Change},
  journal={IEEE Transactions on Pattern Analysis and Machine Intelligence}, 
  title={Flare7K++: Mixing Synthetic and Real Datasets for Nighttime Flare Removal and Beyond}, 
  year={2024},
  volume={46},
  number={11},
  pages={7041-7055},
  doi={10.1109/TPAMI.2024.3406821}}

@article{jiang2021enlightengan,
  title={Enlightengan: Deep light enhancement without paired supervision},
  author={Jiang, Yifan and Gong, Xinyu and Liu, Ding and Cheng, Yu and Fang, Chen and Shen, Xiaohui and Yang, Jianchao and Zhou, Pan and Wang, Zhangyang},
  journal={IEEE transactions on image processing},
  volume={30},
  pages={2340--2349},
  year={2021},
  publisher={IEEE}
}

@article{gui2026brightness,
  title={Brightness-aware synthetic-to-real learning for nighttime hazy image enhancement},
  author={Gui, Jie and Cong, Xiaofeng and Zhang, Yu-Xin and Hou, Junming and Tao, Dacheng},
  journal={IEEE Transactions on Pattern Analysis and Machine Intelligence},
  year={2026},
  publisher={IEEE}
}

@inproceedings{sharma2021nighttime,
  title={Nighttime visibility enhancement by increasing the dynamic range and suppression of light effects},
  author={Sharma, Aashish and Tan, Robby T},
  booktitle={Proceedings of the IEEE/CVF Conference on Computer Vision and Pattern Recognition},
  pages={11977--11986},
  year={2021}
}

@inproceedings{jin2022unsupervised,
  title={Unsupervised night image enhancement: When layer decomposition meets light-effects suppression},
  author={Jin, Yeying and Yang, Wenhan and Tan, Robby T},
  booktitle={European Conference on Computer Vision},
  pages={404--421},
  year={2022},
  organization={Springer}
}

@article{wu2024overall,
  title={For Overall Nighttime Visibility: Integrate Irregular Glow Removal With Glow-Aware Enhancement},
  author={Wu, Wanyu and Wang, Wei and Wang, Zheng and Jiang, Kui and Li, Zhengguo},
  journal={IEEE Transactions on Circuits and Systems for Video Technology},
  year={2024},
  publisher={IEEE}
}

@inproceedings{wu2023generation,
  title     = {From Generation to Suppression: Towards Effective Irregular Glow Removal for Nighttime Visibility Enhancement},
  author    = {Wu, Wanyu and Wang, Wei and Wang, Zheng and Jiang, Kui and Xu, Xin},
  booktitle = {Proceedings of the Thirty-Second International Joint Conference on
               Artificial Intelligence, {IJCAI-23}},
  pages     = {1533--1541},
  year      = {2023},
  month     = {8},
  doi       = {10.24963/ijcai.2023/170},
}

@article{dai2022flare7k,
  title={Flare7k: A phenomenological nighttime flare removal dataset},
  author={Dai, Yuekun and Li, Chongyi and Zhou, Shangchen and Feng, Ruicheng and Loy, Chen Change},
  journal={Advances in Neural Information Processing Systems},
  volume={35},
  pages={3926--3937},
  year={2022}
}

@inproceedings{xie2025learning,
  title={Learning Structural Priors via Laplacian RWKV Diffusion with Light-Effect Dataset for Nighttime Visibility Enhancement},
  author={Xie, Dirui and Hu, Xiaofang and Wei, Zihan and Yang, Zhengqiqi and Jiang, Yanlian and Zhou, Yue},
  booktitle={Proceedings of the 33rd ACM International Conference on Multimedia},
  pages={4590--4599},
  year={2025}
}

@article{leonard2013survey,
  title={A survey of the schr\"odinger problem and some of its connections with optimal transport},
  author={L{\'e}onard, Christian},
  journal={arXiv preprint arXiv:1308.0215},
  year={2013}
}

@InProceedings{pmlr-v202-liu23ai,
  title = 	 {{I}$^2${SB}: Image-to-Image Schrödinger Bridge},
  author =       {Liu, Guan-Horng and Vahdat, Arash and Huang, De-An and Theodorou, Evangelos and Nie, Weili and Anandkumar, Anima},
  booktitle = 	 {Proceedings of the 40th International Conference on Machine Learning},
  pages = 	 {22042--22062},
  year = 	 {2023},
  volume = 	 {202},
  series = 	 {Proceedings of Machine Learning Research},
  month = 	 {23--29 Jul},
  publisher =    {PMLR},
}

@article{gushchin2023entropic,
  title={Entropic neural optimal transport via diffusion processes},
  author={Gushchin, Nikita and Kolesov, Alexander and Korotin, Alexander and Vetrov, Dmitry P and Burnaev, Evgeny},
  journal={Advances in Neural Information Processing Systems},
  volume={36},
  pages={75517--75544},
  year={2023}
}

@article{de2024schrodinger,
  title={Schrodinger bridge flow for unpaired data translation},
  author={De Bortoli, Valentin and Korshunova, Iryna and Mnih, Andriy and Doucet, Arnaud},
  journal={Advances in Neural Information Processing Systems},
  volume={37},
  pages={103384--103441},
  year={2024}
}

@InProceedings{
  kim2023unsb,
  title={Unpaired Image-to-Image Translation via Neural Schrödinger Bridge},
  author={Beomsu Kim and Gihyun Kwon and Kwanyoung Kim and Jong Chul Ye},
  booktitle={ICLR},
  year={2024}
}

@inproceedings{dong2025cunsb,
  title={Cunsb-rfie: Context-aware unpaired neural schr{\"o}dinger bridge in retinal fundus image enhancement},
  author={Dong, Xuanzhao and Vasa, Vamsi Krishna and Zhu, Wenhui and Qiu, Peijie and Chen, Xiwen and Su, Yi and Xiong, Yujian and Yang, Zhangsihao and Chen, Yanxi and Wang, Yalin},
  booktitle={2025 IEEE/CVF Winter Conference on Applications of Computer Vision (WACV)},
  pages={4502--4511},
  year={2025},
  organization={IEEE}
}

@ARTICLE{10814698,
  author={Luo, Yu and Chen, Xuanrong and Ling, Jie and Huang, Chao and Zhou, Wei and Yue, Guanghui},
  journal={IEEE Transactions on Multimedia}, 
  title={Unsupervised Low-Light Image Enhancement With Self-Paced Learning}, 
  year={2025},
  volume={27},
  number={},
  pages={1808-1820},
  doi={10.1109/TMM.2024.3521752}}

@ARTICLE{11125898,
  author={Guo, Ruoyu and Pagnucco, Maurice and Song, Yang},
  journal={IEEE Transactions on Multimedia}, 
  title={Exploring Multi-Feature Relationship in Retinex Decomposition for Low-Light Image Enhancement}, 
  year={2025},
  volume={27},
  number={},
  pages={7271-7284},
  doi={10.1109/TMM.2025.3599099}}

@ARTICLE{11574000,
  author={Fan, Linwei and Cui, Jin and Li, Huiyu and Zhao, Hongyi and Sun, Tian and Zhang, Yongxia and Zhang, Caiming},
  journal={IEEE Transactions on Multimedia}, 
  title={HSV-Driven Illumination-Aware Iterative Network for Unsupervised Low-Light Enhancement}, 
  year={2026},
  volume={28},
  number={},
  pages={11010-11022},
  doi={10.1109/TMM.2026.3705211}}

@inproceedings{lan2025schrodinger,
  title={When Schrodinger Bridge Meets Real-World Image Dehazing with Unpaired Training},
  author={Lan, Yunwei and Cui, Zhigao and Luo, Xin and Liu, Chang and Wang, Nian and Zhang, Menglin and Su, Yanzhao and Liu, Dong},
  booktitle={Proceedings of the IEEE/CVF International Conference on Computer Vision},
  pages={8756--8765},
  year={2025}
}

@inproceedings{wang2023ultra,
  title={Ultra-high-definition low-light image enhancement: A benchmark and transformer-based method},
  author={Wang, Tao and Zhang, Kaihao and Shen, Tianrun and Luo, Wenhan and Stenger, Bjorn and Lu, Tong},
  booktitle={Proceedings of the AAAI conference on artificial intelligence},
  volume={37},
  number={3},
  pages={2654--2662},
  year={2023}
}

@inproceedings{deng2022stytr2,
  title={Stytr2: Image style transfer with transformers},
  author={Deng, Yingying and Tang, Fan and Dong, Weiming and Ma, Chongyang and Pan, Xingjia and Wang, Lei and Xu, Changsheng},
  booktitle={Proceedings of the IEEE/CVF conference on computer vision and pattern recognition},
  pages={11326--11336},
  year={2022}
}

@INPROCEEDINGS{4408899,
  author={Metari, S. and Deschenes, F.},
  booktitle={2007 IEEE 11th International Conference on Computer Vision}, 
  title={A New Convolution Kernel for Atmospheric Point Spread Function Applied to Computer Vision}, 
  year={2007},
  volume={},
  number={},
  pages={1-8},
  doi={10.1109/ICCV.2007.4408899}}

@inproceedings{johnson2016perceptual,
  title={Perceptual losses for real-time style transfer and super-resolution},
  author={Johnson, Justin and Alahi, Alexandre and Fei-Fei, Li},
  booktitle={European conference on computer vision},
  pages={694--711},
  year={2016},
  organization={Springer}
}

@article{li2021learning,
  title={Learning to enhance low-light image via zero-reference deep curve estimation},
  author={Li, Chongyi and Guo, Chunle and Loy, Chen Change},
  journal={IEEE transactions on pattern analysis and machine intelligence},
  volume={44},
  number={8},
  pages={4225--4238},
  year={2021},
  publisher={IEEE}
}

@inproceedings{shi2024zero,
  title={ZERO-IG: Zero-shot illumination-guided joint denoising and adaptive enhancement for low-light images},
  author={Shi, Yiqi and Liu, Duo and Zhang, Liguo and Tian, Ye and Xia, Xuezhi and Fu, Xiaojing},
  booktitle={Proceedings of the IEEE/CVF conference on computer vision and pattern recognition},
  pages={3015--3024},
  year={2024}
}

@inproceedings{zhang2020nighttime,
  title={Nighttime dehazing with a synthetic benchmark},
  author={Zhang, Jing and Cao, Yang and Zha, Zheng-Jun and Tao, Dacheng},
  booktitle={Proceedings of the 28th ACM international conference on multimedia},
  pages={2355--2363},
  year={2020}
}

@article{zhang2019root,
  title={Root mean square layer normalization},
  author={Zhang, Biao and Sennrich, Rico},
  journal={Advances in neural information processing systems},
  volume={32},
  year={2019}
}

@article{wang2022variational,
  title={Variational single nighttime image haze removal with a gray haze-line prior},
  author={Wang, Wenhui and Wang, Anna and Liu, Chen},
  journal={IEEE Transactions on Image Processing},
  volume={31},
  pages={1349--1363},
  year={2022},
  publisher={IEEE}
}

@InProceedings{Ershov_2025_CVPR,
    author    = {Ershov, Egor and Korchagin, Sergey and Khalin, Aleksei and Panshin, Artyom and et al. },
    title     = {NTIRE 2025 Challenge on Night Photography Rendering},
    booktitle = {Proceedings of the IEEE/CVF Conference on Computer Vision and Pattern Recognition (CVPR) Workshops},
    month     = {June},
    year      = {2025},
    pages     = {1514-1524}
}

@inproceedings{jin2023enhancing,
  title={Enhancing visibility in nighttime haze images using guided apsf and gradient adaptive convolution},
  author={Jin, Yeying and Lin, Beibei and Yan, Wending and Yuan, Yuan and Ye, Wei and Tan, Robby T},
  booktitle={Proceedings of the 31st ACM international conference on multimedia},
  pages={2446--2457},
  year={2023}
}

@article{song2023vision,
  title={Vision transformers for single image dehazing},
  author={Song, Yuda and He, Zhuqing and Qian, Hui and Du, Xin},
  journal={IEEE Transactions on Image Processing},
  volume={32},
  pages={1927--1941},
  year={2023},
  publisher={IEEE}
}

@inproceedings{cong2024semi,
  title={A semi-supervised nighttime dehazing baseline with spatial-frequency aware and realistic brightness constraint},
  author={Cong, Xiaofeng and Gui, Jie and Zhang, Jing and Hou, Junming and Shen, Hao},
  booktitle={Proceedings of the IEEE/CVF Conference on Computer Vision and Pattern Recognition},
  pages={2631--2640},
  year={2024}
}

@inproceedings{zhu2017unpaired,
  title={Unpaired image-to-image translation using cycle-consistent adversarial networks},
  author={Zhu, Jun-Yan and Park, Taesung and Isola, Phillip and Efros, Alexei A},
  booktitle={Proceedings of the IEEE international conference on computer vision},
  pages={2223--2232},
  year={2017}
}

@article{thurstone1927law,
  author  = {Thurstone, L. L.},
  title   = {A Law of Comparative Judgment},
  journal = {Psychological Review},
  volume  = {34},
  number  = {4},
  pages   = {273--286},
  year    = {1927}
}

@inproceedings{zhang2023liqe,  
  title={Blind Image Quality Assessment via Vision-Language Correspondence: A Multitask Learning Perspective},  
  author={Zhang, Weixia and Zhai, Guangtao and Wei, Ying and Yang, Xiaokang and Ma, Kede},  
  booktitle={IEEE Conference on Computer Vision and Pattern Recognition},  
  pages={14071--14081},
  year={2023}
}

@article{chu2017cyclegan,
  title={Cyclegan, a master of steganography},
  author={Chu, Casey and Zhmoginov, Andrey and Sandler, Mark},
  journal={arXiv preprint arXiv:1712.02950},
  year={2017}
}

@inproceedings{wang2023clipiqa,
  title     = {Exploring {CLIP} for Assessing the Look and Feel of Images},
  author    = {Wang, Jie and Chan, Kelvin C. and Loy, Chen Change},
  booktitle = {Proceedings of the AAAI Conference on Artificial Intelligence},
  volume    = {37},
  number    = {2},
  pages     = {2555--2563},
  year      = {2023}
}

@inproceedings{
feng2026seeing,
title={Seeing Across Views: Benchmarking Spatial Reasoning of Vision-Language Models in Robotic Scenes},
author={ZhiYuan Feng and Zhaolu Kang and Qijie Wang and Zhiying Du and Jiongrui Yan and Shi Shubin and Chengbo Yuan and Huizhi Liang and Yu Deng and Qixiu Li and Rushuai Yang and Ruichuan An and Leqi Zheng and Weijie Wang and Shuang Chen and Sicheng Xu and Yaobo Liang and Jiaolong Yang and Baining Guo},
booktitle={The Fourteenth International Conference on Learning Representations},
year={2026},
}

@techreport{openai2025gpt5,
  author       = {{OpenAI}},
  title        = {{GPT-5.6 System Card}},
  institution  = {OpenAI},
  year         = {2026},
  url          = {https://deploymentsafety.openai.com/gpt-5-6/gpt-5-6.pdf},
  note         = {Accessed: July 09, 2026}
}

@INPROCEEDINGS{5692551,
  author={Kovesi, Peter},
  booktitle={2010 International Conference on Digital Image Computing: Techniques and Applications}, 
  title={Fast Almost-Gaussian Filtering}, 
  year={2010},
  volume={},
  number={},
  pages={121-125},
  doi={10.1109/DICTA.2010.30}}

@ARTICLE{9548827,
  author={Ou, Fu-Zhao and Wang, Yuan-Gen and Li, Jin and Zhu, Guopu and Kwong, Sam},
  journal={IEEE Transactions on Multimedia}, 
  title={A Novel Rank Learning Based No-Reference Image Quality Assessment Method}, 
  year={2022},
  volume={24},
  number={},
  pages={4197-4211},
  doi={10.1109/TMM.2021.3114551}}

@INPROCEEDINGS{1211417,
  author={Narasimhan, S.G. and Nayar, S.K.},
  booktitle={2003 IEEE Computer Society Conference on Computer Vision and Pattern Recognition, 2003. Proceedings.}, 
  title={Shedding light on the weather}, 
  year={2003},
  volume={1},
  number={},
  pages={I-I},
  doi={10.1109/CVPR.2003.1211417}}

@inproceedings{wu2024qalign,
  title     = {Q-Align: Teaching {LMM}s for Visual Scoring via Discrete Text-Defined Levels},
  author    = {Wu, Haoning and Zhang, Zicheng and Zhang, Weixia and Chen, Chaofeng and
               Liao, Liang and Li, Chunyi and Gao, Yixuan and Wang, Annan and Zhang, Erli and
               Sun, Wenxiu and Yan, Qiong and Min, Xiongkuo and Zhai, Guangtao and Lin, Weisi},
  booktitle = {Proceedings of the 41st International Conference on Machine Learning (ICML)},
  year      = {2024}
}

@inproceedings{yang2022maniqa,
  title={MANIQA: Multi-dimension Attention Network for No-Reference Image Quality Assessment},
  author={Yang, Sidi and Wu, Tianhe and Shi, Shuwei and Lao, Shanshan and Gong, Yuan and Cao, Mingdeng and Wang, Jiahao and Yang, Yujiu},
  booktitle={Proceedings of the IEEE/CVF Conference on Computer Vision and Pattern Recognition},
  pages={1191--1200},
  year={2022}
}

\newpage

 




\vfill

\end{document}